\documentclass{article}

\usepackage{microtype}
\usepackage{graphicx}
\usepackage{booktabs} %

\usepackage{hyperref}

\usepackage[accepted]{icml2023}

\usepackage{amsmath}
\usepackage{amssymb}
\usepackage{mathtools}
\usepackage{amsthm}

\usepackage{rotating}

\usepackage[capitalize,noabbrev]{cleveref}

\theoremstyle{plain}

\theoremstyle{definition}

\theoremstyle{remark}

\usepackage[textsize=tiny]{todonotes}

\usepackage{lipsum}
\usepackage{natbib}
\usepackage{amsfonts}
\usepackage{graphicx}
\usepackage{subcaption}
\usepackage{environ}
\usepackage{xcolor}
\usepackage{tikz}
\usepackage{pgfplots}
\usepackage{pgfplotstable}
\usepackage{etoolbox}
\usepackage{calc}
\usepackage{acro}
\usepackage{hyperref}

\pgfplotsset{compat=1.9}

\pgfplotsset{
    discard if/.style 2 args={
        x filter/.append code={
            \edef\tempa{\thisrow{#1}}
            \edef\tempb{#2}
            \ifx\tempa\tempb
            
            \fi
        }
    },
    discard if not/.style 2 args={
        x filter/.append code={
            \edef\tempa{\thisrow{#1}}
            \edef\tempb{#2}
            \ifx\tempa\tempb
            \else
            
            \fi
        }
    }
}

\newcommand{\maxlength}{\ell_{\mathrm{max}}}

\newcommand{\iidsim}{\overset{\text{i.i.d.}}\sim}

\DeclareMathOperator{\distUnifOp}{Unif}
\newcommand{\distUnif}[1]{\distUnifOp\left(#1\right)}

\DeclareMathOperator{\probMathOp}{Pr}
\newcommand{\prob}[1]{\probMathOp\left(#1\right)}
\newcommand{\cprob}[2]{\prob{#1\middle|#2}}
\newcommand{\reals}{\mathbb{R}}

\newcommand{\samplespace}{\mathcal{X}}
\newcommand{\alphabet}{\mathcal{S}}
\newcommand{\rkhs}{\mathcal{H}}

\newcommand{\Verts}[1]{\lVert#1\rVert}
\newcommand{\verts}[1]{\lvert#1\rvert}

\newcommand{\steinOp}{\mathcal{A}}

\DeclareMathOperator{\ksdOp}{KSD}

\newcommand{\ksdmod}[2]{\ksdOp\bigl[#1\Vert#2\bigr]}

\newcommand{\ksdsqmod}[2]{\ksdOp^2\bigl[#1\Vert#2\bigr]}

\newcommand{\ksdustat}[3]{U_{#1, #2}\left(#3\right)}
\newcommand{\ksdwildustat}[4]{U_{#1, #2}^{#4}\left(#3\right)}

\newcommand{\neigh}[1]{\partial #1}
\newcommand{\balfunNoArg}{g}
\newcommand{\balfun}[1]{g\left(#1\right)}
\DeclareMathOperator{\delAction}{del}
\DeclareMathOperator{\insAction}{ins}

\DeclareMathOperator{\repAction}{rep}

\newcommand{\dit}{P}
\newcommand{\pmft}{p}
\newcommand{\dis}{Q}

\newcommand{\disSamples}{\left\{Y^{(l)}\right\}_{l=1}^m}

\newcommand{\disSampleCount}{m}

\newcommand{\of}[1]{\left(#1\right)}

\newcommand{\vertices}{\mathcal{V}}
\newcommand{\edges}{\mathcal{E}}

\DeclareAcronym{mmd}{
    short=MMD,
    long=maximum mean discrepancy,
    long-plural-form=maximum mean discrepancies,
}
\DeclareAcronym{ksd}{
    short=KSD,
    long=kernel Stein discrepancy,
    long-plural-form=kernel Stein discrepancies,
}
\DeclareAcronym{ipm}{
    short=IPM,
    long=integral probability metric,
}
\DeclareAcronym{rkhs}{
    short=RKHS,
    long=reproducing kernel Hilbert space,
}
\DeclareAcronym{pd}{
    short=PD,
    long=positive definite,
}
\DeclareAcronym{dtmc}{
    short=DTMC,
    long=discrete time Markov chain,
}
\DeclareAcronym{ctmc}{
    short=CTMC,
    long=continuous time Markov chain,
}
\DeclareAcronym{mc}{
    short=MC,
    long=Markov chain,
}
\DeclareAcronym{lr}{
    short=LR,
    long=likelihood ratio,
}
\DeclareAcronym{gof}{
    short=GoF,
    long=goodness-of-fit,
}

\definecolor{settwo1}{rgb}{0.4,0.76078431372549016,0.6470588235294118}
\definecolor{settwo2}{rgb}{0.9882352941176471,0.55294117647058827,0.3843137254901961}
\definecolor{settwo3}{rgb}{0.55294117647058827,0.62745098039215685,0.79607843137254897}
\definecolor{settwo4}{rgb}{0.90588235294117647,0.54117647058823526,0.76470588235294112}
\definecolor{settwo5}{rgb}{0.65098039215686276,0.84705882352941175,0.32941176470588235}
\definecolor{settwo6}{rgb}{1.0,0.85098039215686272,0.18431372549019609}
\definecolor{settwo7}{rgb}{0.89803921568627454,0.7686274509803922,0.58039215686274515}
\definecolor{settwo8}{rgb}{0.70196078431372544,0.70196078431372544,0.70196078431372544}

\pgfplotscreateplotcyclelist{barcyclelist}{
        {draw=none,fill=settwo1,mark=none},
        {draw=none,fill=settwo2,mark=none},
        {draw=none,fill=settwo3,mark=none},
        {draw=none,fill=settwo4,mark=none},
        {draw=none,fill=settwo5,mark=none},
        {draw=none,fill=settwo6,mark=none},
        {draw=none,fill=settwo7,mark=none},
        {draw=none,fill=settwo8,mark=none},
}

\pgfplotscreateplotcyclelist{linecyclelist}{
        {draw=settwo1,mark=+},
        {draw=settwo2,mark=x},
        {draw=settwo3,mark=|},
        {draw=settwo4,mark=triangle},
        {draw=settwo5,mark=square},
        {draw=settwo6,mark=pentagon},
        {draw=settwo7,mark=asterisk},
        {draw=settwo8,mark=o},
}

\DeclareMathOperator{\simplexOp}{Sim}
\newcommand{\simplex}[1]{\simplexOp_{#1}}

\begin{document}

\twocolumn[
\icmltitle{A Kernel Stein Test of Goodness of Fit for Sequential Models}

\icmlsetsymbol{equal}{*}

\begin{icmlauthorlist}
\icmlauthor{Jerome Baum}{equal,uclcs}
\icmlauthor{Heishiro Kanagawa}{equal,gatsby,newcastle}
\icmlauthor{Arthur Gretton}{gatsby}
\end{icmlauthorlist}

\icmlaffiliation{uclcs}{Department of Computer Science, University College London, UK}
\icmlaffiliation{gatsby}{Gatsby Computational Neuroscience Unit, UCL}
\icmlaffiliation{newcastle}{School of Mathematics, Statistics and Physics, Newcastle University, UK}

\icmlcorrespondingauthor{Jerome Baum}{jerome@jeromebaum.com}
\icmlcorrespondingauthor{Heishiro Kanagawa}{heishiro.kanagawa@gmail.com}

\icmlkeywords{Machine Learning, ICML, Kernel Stein Discrepancy, Goodness of Fit}

\vskip 0.3in
]

\printAffiliationsAndNotice{\icmlEqualContribution} %

\begin{abstract}
    We propose a goodness-of-fit measure for probability densities modeling observations with varying dimensionality, such as text documents of differing lengths or variable-length sequences.
    The proposed measure is an instance of the kernel Stein discrepancy (KSD), which has been used to construct goodness-of-fit tests for unnormalized densities.
    The KSD is defined by its Stein operator: current operators used in testing apply to fixed-dimensional spaces. 
    As our main contribution, we extend the KSD to the variable-dimension setting by identifying appropriate Stein operators, and propose a novel KSD goodness-of-fit test.
    As with the previous variants, the proposed KSD does not require the density to be normalized, allowing the evaluation of a large class of models.
    Our test is shown to perform well in practice on discrete sequential data benchmarks.
\end{abstract}

    \section{Introduction}\label{sec:intro}

    This paper addresses the problem of evaluating a probabilistic model for observations with variable dimensionality.
This problem commonly arises in sequential data analysis, where sequences of different lengths are observed, and generative models (e.g., Markov models) are used to draw inferences.
Our problem setting also concerns a scenario where a data point is a collection of observations that may not be sequentially ordered, as in the topic modeling of text documents.
Our task is formalized as follows: given a sample $\{X_i\}_{i=1}^n$ from a distribution $\dis$ on a sample space $\samplespace$ (e.g., the set of all possible sequences),
we aim to quantify the discrepancy of distribution $\dit$ modeling $\dis$.

One approach to this problem is generating samples from the model and computing a sample-based discrepancy measure, such as the maximum mean discrepancy (MMD)~\citep{gretton_kernel_2012, LloGha15}.
A disadvantage of this approach is that it potentially requires many samples to see the  departure of the model from the data.
For example, if a sequence model misspecifies a particular transition from a given history, we would need to generate sequences sharing this history to observe the disparity. 
Unfortunately, the MMD approach becomes less efficient as the state space enlarges or the length of the history grows, necessitating repeated sampling.
This observation motivates using a measure that exploits information provided by the model, such as dependence relations between different time points.

A model-dependent measure may be derived based on Stein's method~\citep{stein_bound_1972}, a technique from probability theory developed originally to obtain explicit rates of convergence to normality.
The key construct from Stein's method is a distribution-specific operator called a Stein operator that modifies a function to have zero expectation under the distribution.
A model-specific Stein operator $\steinOp_{\dit}$ may be defined to construct a zero-expectation function $\steinOp_{\dit}f$; its expectation $\mathbb{E}_{X\sim\dis}[\steinOp_{\dit}f(X)]$ under the sample distribution serves as a discrepancy measure, since a non-zero expectation indicates the deviation from the model.
One may generalize this idea to a family of functions $\mathcal{F}$, and the resulting worst-case summary $\sup_{f\in\mathcal{F}} \lvert\mathbb{E}_{X\sim\dis}[\steinOp_{\dit}f(X)]\rvert$ is called a Stein discrepancy, proposed by \citet{gorham_measuring_2015}.
An appropriate choice of the operator and the function class yields a computable Stein discrepancy. 
This paper focuses on an instance called the kernel Stein discrepancy (KSD)~\citep{Oates_2016, chwialkowski_kernel_2016, liu_kernelized_2016, GorMac2017}, where functions from a reproducing kernel Hilbert space (RKHS) are used.

The KSD is a versatile framework for designing goodness-of-fit measures.
Given a Stein operator and a reproducing kernel, the KSD is expressed by an expectation of a Stein-modified kernel, leading to tractable estimators only involving sample evaluations of the modified kernel.
This feature allows us to use a wealth of kernels from the literature: by designing a Stein operator, one can adapt the kernel to the model and define a bespoke discrepancy measure. 
Based on the zero-mean kernel theory by~\citet{Oates_2016}, \citet{chwialkowski_kernel_2016} and \citet{liu_kernelized_2016} originated this line of work, where they combined the Langevin Stein operator proposed by \citet{gorham_measuring_2015} with an RKHS; remarkably, the resulting (Langevin) KSD is computable for densities with unknown normalizing constants.
There have been numerous extensions to models in other data domains:
categorical data~\citep{yang_goodness--fit_2018}, point-pattern data~\citep{yang_2019}, censored data~\citep{fernandez_kernelized_2020}, directional data~\citep{xu_2020}, and functional data~\citep{Wynne2022}.
We refer the reader to~\citet{anastasiou_steins_2021} for a review on the KSD's applications outside goodness-of-fit testing.

A limitation of the preceding Stein works is that they require the model distribution to be defined on a fixed-dimension space.
For example, \citet{kanagawa_kernel_2019} evaluate latent Dirichlet allocation models~\citep{BleNgJor03} for text documents assuming a fixed document length, an assumption unlikely to hold in practice.
Relevant work has been accomplished by~\citet{Wynne2022}, where they propose a KSD for distributions on an (infinite-dimensional) Hilbert space, and treat continuous-time models, such as Brownian motions and stochastic differential equation models.
While their test applies to sequential models, our target setting is different, as we focus on discrete-time models such as Markov chains, and do not require a density with respect to a Gaussian measure.

In the present work, we extend the KSD framework to the variable-dimension setting.
Specifically, we identify a Stein operator for distributions on the set of all univariate sequences of finite alphabets. 
This is to our knowledge the first such operator, and it may of independent interest. 
Our Stein operator builds on the class of Zanella-Stein operators recently proposed by~\citet{hodgkinson_reproducing_2020}, which are derived from a Markov jump process having the target distribution as invariant measure;
we review the Zanella-Stein operator in Section~\ref{sec:background}.
In Section \ref{sec:method}, we derive a new Stein operator using a Markov process admitting transitions between sets of different sizes.
Based on our Stein operator and the associated KSD, we propose a novel goodness-of-fit test for sequential models.
As in previous variants, the proposed KSD does not require the model to be normalized, allowing the evaluation of \emph{intractable models}, including Markov random fields (\cref{sec:mrf-experiment}) and conditional generative models (\cref{sec:additional-exp-conditional}).
As the proposed operator involves a number of tunable parameters, as our second contribution, we offer guidance on how to select these based on empirical power analysis.

    \section{Background}\label{sec:background}

    All Stein discrepancies are build around a particular Stein operator. 
In this section, we recall a class of operators, the Zanella-Stein operators, 
and describe the challenges in constructing operators of this class in the sequential setting. 
We also review the associated kernel Stein discrepancy (KSD), a goodness-of-fit measure that leverages a given Stein operator to construct a test statistic. 

\paragraph{Zanella-Stein Operator}
The Zanella-Stein (ZS) operators are a class of Stein operators for distributions on a countable set $\samplespace$, proposed by~\citet{hodgkinson_reproducing_2020} following the generator method of~\citet{barbour_steins_1988}.
In the following, we assume that the model $\dit$ has probability mass function $\pmft$ positive everywhere in $\samplespace$.
For a real-valued function $f$ on $\samplespace$, a ZS operator $\steinOp_{\dit}$ is defined by
\begin{align}
    \of{\steinOp_{\dit} f}(x)
    &\coloneqq
    \sum_{y\in\neigh{x}} \balfun{\frac{\pmft(y)}{\pmft(x)}} \left(f(y) - f(x)\right), \label{eqn:steinop}
\end{align}
where $\neigh{x}\subset \samplespace$ is a set of \emph{neighborhood} points, and $\balfunNoArg:(0,\infty)\to (0,\infty) $ is a \emph{balancing function}, which must satisfy $\balfun{t}=t\balfun{1/t}$ for $t>0$.
Several choices of $\balfunNoArg$ are known in the literature: the minimum probability flow operator discussed in~\citep{BarpBriolDuncanEtAl2019Minimum} uses $\balfun{t}=\sqrt{t}$; the choice $\balfun{t}=t/(1+t)$ is known as the Barker Stein operator~\citep{Barker_1965, shi_gradient_2022}.

A Zanella-Stein operator is the infinitesimal generator
of a Markov jump process called a Zanella process~\citep{zanella_informed_2019,power_accelerated_2019} designed to have the target distribution $\dit$ as an invariant distribution.
The process is based on the idea of \emph{locally informed proposals}:
these jump from a given point $x$ to any of its neighbors $y$, at a rate $\balfun{\pmft(y)/\pmft(x)}$ so that
 detailed balance is satisfied for $\dit$,
ensuring that $\dit$ is an invariant distribution.
For the operator to characterize $\dit$, we must specify our notion of neighborhood. 
Let $(\vertices,\edges)$ be
the directed graph with vertices $\vertices=\samplespace$ and edges $\edges=\{(x, y) | y\in\neigh{x}\}$ induced by the chosen neighborhood structure.
The graph $(\vertices,\edges)$ must have the following properties:
\begin{enumerate}
    \item Symmetry: $(x,y)\in\edges \Leftrightarrow (y,x)\in\edges$.
    \item Strong connectivity: the transitive closure of $\edges$ is $\samplespace\times\samplespace$,
    so that there is a path from every point to every other point.
\end{enumerate}
In fact, symmetry alone implies that
$\dit$ is in the set of invariant distributions;
strong connectivity is required to ensure that
$\dit$ is the unique invariant distribution, as otherwise the corresponding Stein discrepancy cannot distinguish $\dit$ from other distributions even for a rich test function class.
In \Citet{hodgkinson_reproducing_2020}, the aperiodicity condition is additionally required. 
For a pure jump-type process, we only need the irreducibility of the process \citep[][Theorem 13.12]{Kallenberg_2021}, and hence we may dispense with this condition. 
Note that we can conveniently choose a sparse neighborhood to accelerate the computation, although this may result in reduced sensitivity of the discrepancy.
In Section~\ref{sec:experiment}, we discuss in more detail the tradeoff between computational cost and test power.

\paragraph{Challenges in the Variable-Dimension Setting}
The generator method (as in the ZS operator above) allows us to obtain an operator for any state space with an appropriate Markov process.
Indeed, the Zanella process is not the only allowable choice: other Markov chains or processes have also been considered for discrete spaces, including birth-death processes~\citep{shi_gradient_2022}, the Glauber dynamics Markov chain~\citep{ReiRos19, BreNag19} 
\citep[we refer the reader to][for a review]{shi_gradient_2022}.
Defining a Markov process is often a challenge, however; prior work has thus only considered processes in fixed-dimensional spaces, 
and has not dealt with sequential models such as those considered here. 
The ZS operators are particularly attractive in our setting, since they may be defined on any discrete set. 
The main challenge in deriving an operator in this class is the construction of an appropriate neighborhood for 
models of variable-length sequences, which is highly nontrivial. 
Our contribution is to develop an effective neighborhood definition for the sequential setting, 
which satisfies the three requirement outlined above, and yields an associated test that has good statistical power against alternatives.

\paragraph{Kernel Stein Discrepancy}
We next review the construction of a test statistic from the Stein operator. 
A computable discrepancy measure may be defined using a reproducing kernel Hilbert space (RKHS)~\citep{aronszajn_theory_1950, steinwart_support_2008}.
Following~\citep{Oates_2016, chwialkowski_kernel_2016, liu_kernelized_2016, GorMac2017},
we define the kernel Stein discrepancy (KSD) associated with our (yet to be specified) Zanella-Stein operator as follows:
\begin{align}
    \ksdmod{\dis}{\dit} \coloneqq \sup_{\Verts{f}_{\rkhs}\leq 1}  \bigl|\mathbb{E}_{X\sim\dis} [\steinOp_{\dit} (f)(X)]\bigr|,
    \label{eqn:ksd-def}
\end{align}
where $\rkhs$ is the RKHS of real-valued functions on $\samplespace$ corresponding to a positive definite kernel $k: \samplespace \times \samplespace \to \reals$ with the natural norm $\Verts{f}_{\rkhs} = \sqrt{\langle f, f \rangle_{\rkhs}}$ defined by the inner product $\langle \cdot, \cdot \rangle_{\rkhs}$.
By the vanishing property $\mathbb{E}_{Y\sim \dit}[\steinOp_{\dit}f(Y)]=0$,  the KSD is zero if $\dis=\dit$. The other direction requires further assumptions on the operator and the RKHS.
According to~\citet[Proposition 4]{hodgkinson_reproducing_2020}, when the corresponding Zanella process is exponentially ergodic and the RKHS is $C_0$-universal, we have $\ksdmod{\dis}{\dit}=0$ only if $\dit = \dis$.
For a finite state space $\samplespace$, exponential ergodicity is satisfied by any irreducible Markov jump process.
The $C_0$-universality is equivalent to the integrally strict positive definiteness \citep{SriFukLan2011}; 
in the particular case $\verts{\samplespace}<\infty$, this condition states that the Gram matrix defined over all the configurations on $\samplespace$ is strictly positive definite.

The use of an RKHS yields a closed-form expression of the KSD~\citep[][see Proposition 1 and the paragraph following the proof]{hodgkinson_reproducing_2020}:
\begin{align}
    \ksdsqmod{\dis}{\dit} = \mathbb{E}_{X, X'\sim \dis\otimes\dis} [h_{\pmft} (X, X')],
    \label{eqn:ksdsq-def}
\end{align}
where $X$, $X'$ are i.i.d. random variables with law $\dis$, provided $\mathbb{E}_{X\sim\dis}[h_{\pmft} (X, X)^{1/2}]<\infty$.
The proof of this result follows as in~\citep[][Proposition 2]{GorMac2017}, noting that the inner product between $f$ and $\steinOp_{\dit}k(x, \cdot)$ \emph{reproduces} $\steinOp_{\dit}f(x)$ for any $f \in \rkhs$.
The function $h_{\pmft}$ is called a \emph{Stein kernel}, defined by
\begin{align*}
    \begin{aligned}
    &h_{p}(x,y) \\
    &=\sum_{\nu\in{\cal N}_{x}}\sum_{\tilde{\nu}\in{\cal N}_{y}} \balfunNoArg_{\nu}(x)\balfunNoArg_{\tilde{\nu}}(y)\left\{ k(\nu(x),\tilde{\nu}(y))+k(x,y)\right.\\
        & \hphantom{=\sum_{\nu\in{\cal N}_{x}}\sum_{\tilde{\nu}\in{\cal N}_{y}} \balfunNoArg_{\nu}(x)\balfunNoArg_{\tilde{\nu}}(y)}\left.\quad -k(x,\tilde{\nu}(y))-k(\nu(x),y)\right\}
    \end{aligned},
    \label{eqn:stein-kernel-def}
\end{align*}
where we identify each point in the neighborhood $\neigh{x}$ as a mapping from $x$ to itself, and $\balfunNoArg_{\nu}(x) = \balfunNoArg\{\pmft(\nu(x))/\pmft(x)\}$.

Given a sample $\{X_i\}_{i=1}^n \iidsim \dis$, the squared KSD expression \eqref{eqn:ksdsq-def} admits a simple unbiased estimator
\begin{equation}
\ksdustat{n}{\dit}{\{X_i\}_{i=1}^n} \coloneqq \frac{1}{n(n-1) }\sum_{i \neq j}^n h_{\pmft} (X_i, X_j),
    \label{eqn:ustat}
\end{equation}
which is a U-statistic~\citep{hoeffding_class_1948}.
By the zero-mean property of the Stein kernel, the U-statistic is degenerate if $\dis = \dit$~\citep{chwialkowski_kernel_2016, liu_kernelized_2016}.
In this case, the scaled statistic $n\ksdustat{n}{\dit}{ \{X_i\}_{i=1}^n   }$ asymptotically follows the law of $\sum_{j=1}^\infty \lambda_j(Z_j^2-1)$, where $Z_j$ are i.i.d. standard normal variables, and $\lambda_j$ are eigenvalues given by the eigenvalue problem  $\sum_{x' \in \samplespace}h_p(x,x')a(x')\pmft(x') = \lambda a(x)$ with $\sum_{x\in\samplespace}a(x)^2p(x)<\infty$~\citep[see, e.g.,][Section 5.5]{Ser2009}.
One of our proposed tests in Section~\ref{sec:method} simulates this distribution using a wild bootstrap procedure.

\paragraph{Kernels for Sequences}
The performance of the test developed in the next section depends on the choice of the reproducing kernel. 
As mentioned above, the $C_0$-universality is a condition that guides kernel selection since it renders the test consistent (along with the exponential ergodicity). 
A trivial example of $C_0$-universal kernels is the Dirac kernel that outputs $1$ if two inputs are identical, and $0$ otherwise; but this kernel might not be useful in practice, 
since all data points in the corresponding feature space are orthogonal, and provide no information about each other.
As this example shows, kernels need not be universal to provide powerful tests, as long as they encode relevant features to the setting where they are used.
The literature on sequence kernels is well-established~\citep[see, e.g.,][for a recent account]{kiraly_kernels_2019}, 
and one could use existing kernels, such as global alignment kernels~\citep[][]{Cuturi_2007,Cuturi2011} and string kernels~\citep{haussler_convolution_1999,lodhi_text_2002,Leslie2004}. 
Alternatively, one may also define a kernel by the inner product of 
explicit features (e.g., neural network embedding of sequences) instead of known kernels.

    \section{Testing Sequential Models}\label{sec:method}
    We now address the design of a Stein operator for the variable-dimension setting.
We begin by formally defining the sample space $\samplespace$. 
Let $\alphabet$ be a nonempty finite set of symbols.
For an integer $\ell \geq 1$,
we denote by $\alphabet^{(\ell)}$ the set of all length-$\ell$ sequences formed by symbols in $\alphabet$; i.e., $\alphabet^{(\ell)}=\prod_{j=1}^{\ell}\alphabet$.
Then, the sample space $\samplespace$ is given by the set of all sequences $\bigsqcup_{\ell=1}^{\maxlength} \alphabet^{(\ell)}$,
where $\bigsqcup$ denotes the disjoint union, and $\maxlength$ is $\infty$ or a finite positive integer.
In this setting, a random sample in $\samplespace$ is a sequence whose length is randomly determined.

\begin{figure*}
    \begin{minipage}{0.95\columnwidth}
        \begin{algorithm}[H]
            \caption{Parametric bootstrap test}
            \label{alg:param-bootstrap}
            \begin{algorithmic}[1]
                \REQUIRE $\dit$; a target distribution on $\samplespace$.
                \REQUIRE $D_n=\{X_1,\dots, X_n\}$; data.
                \REQUIRE $\alpha$; the desired level of the test.
                    \FOR{$b=1,\ldots,B$}
                        \STATE Sample $D_n^{(b)}=\{X^{(b)}_1, \dots, X^{(b)}_n\} \iidsim \dit$
                        \STATE $\Delta_{b}\gets \ksdustat{n}{\dit}{D_n^{(b)}}$
                    \ENDFOR
                    \STATE $\hat{t}_{\alpha} \gets (1-\alpha)$-quantile of $\{\Delta_1, \dots, \Delta_B\}$
                    \IF{$\ksdustat{n}{\dit}{D_n} > \hat{t}_{\alpha}$}
                        \STATE Reject $H_0$
                    \ENDIF

            \end{algorithmic}
        \end{algorithm}
    \end{minipage}
    \hfill
    \begin{minipage}{0.95\columnwidth}
        \begin{algorithm}[H]
            \caption{Wild bootstrap test}
            \label{alg:wild-bootstrap}
            \begin{algorithmic}[1]
                \REQUIRE $\dit$; a target distribution on $\samplespace$.
                \REQUIRE $D_n=\{X_1,\dots, X_n\}$; data.
                \REQUIRE $\alpha$; the desired level of the test.
                    \FOR{$b=1,\ldots,B$}
                        \STATE Draw $W_b\sim\mathrm{Multi}(n; 1/n,\dots,1/n)$
                        \STATE $\Delta_{b}\gets \ksdwildustat{n}{\dit}{D_n}{W_b}$ 
                    \ENDFOR
                    \STATE $\hat{t}_{\alpha} \gets (1-\alpha)$-quantile of $\{\Delta_1, \dots, \Delta_B\}$
                    \IF{$\ksdustat{n}{\dit}{D_n} > \hat{t}_{\alpha}$}
                        \STATE Reject $H_0$
                    \ENDIF
            \end{algorithmic}
        \end{algorithm}
    \end{minipage}
\end{figure*}

As noted in Section~\ref{sec:background}, to obtain a KSD in this setting, we need to design a neighborhood structure which specifies our ZS operator.
We first introduce a simple structure for sequences, which edits only the end of the string, to illustrate the required connectivity principles.
We then describe a more advanced neighborhood structure, allowing for greater modifications, which will be the foundation of our tests.

\paragraph {Neighborhood Choice}
For the Stein operator to uniquely characterize the model, we require the strong connectivity of the induced graph.
We achieve this by introducing \emph{inter- and intra-length state transitions}.
Specifically, for a length-$\ell$ sequence $x_{(1:\ell)}=(x_1,\dots,x_\ell)\in \alphabet^{(l)}$, we propose the following neighborhood:

\begin{align}
    \neigh{x_{(1:\ell)}} = \mathcal{I}_{x}  \cup \{x_{(1:\ell-1)}\} \cup \mathcal{R}_{x}
    \label{eqn:neigh-consistent-vanilla}
\end{align}

with
\begin{align*}
    & \mathcal{I}_{x} = \{(x_1, \dots, x_{\ell}, s): s\in \alphabet\}, \\ 
    & \mathcal{R}_{x} = \{(x_1, \dots, x_{l-1}, s): s\in \mathcal{N}_{\mathrm{rep}}(x_{\ell}) \},
\end{align*}
where $\mathcal{N}_{\mathrm{rep}}(s)$ denotes a neighborhood chosen for a symbol $s\in\alphabet$;
this is chosen such that the induced graph is strongly connected.
The first two sets in \eqref{eqn:neigh-consistent-vanilla} represent the \emph{inter-length transitions}:
Each point of $\mathcal{I}_{x}$ corresponds to adding a symbol to the end of the given sequence $x_{(1:\ell)}$,
whereas $ \{x_{(1:l-1)}\}$ stands for the deletion of the last symbol and shortening the sequence.
The third set $\mathcal{R}_{x} $ represents \emph{intra-length transitions}, where the last symbol of the sequence is replaced.
A trivial choice for satisfying the strong connectivity is using the whole alphabet set $\alphabet$ 
for $\mathcal{N}_{\mathrm{rep}}(s)$ for each $s\in \alphabet$.
This approach may be permitted when the alphabet size is small.
One might otherwise want to consider using smaller sets to reduce the computational cost.
For example, we can use a smaller $\mathcal{N}_{\mathrm{rep}}(s)$, provided that the graph (with vertices $\alphabet$) induced by the choice satisfies the three conditions required in Section~\ref{sec:background};
e.g., as in \citep{yang_goodness--fit_2018}, we may introduce a cyclic order $\{0, \dots, \verts{\alphabet}-1\}$ to $\alphabet$, and take two adjacent points $\mathcal{N}_{\mathrm{rep}}(s) = \{s+1, s-1\}$, where $s\pm1$ denotes $ (s\pm1)\ \mathrm{mod}\ \verts{\alphabet}-1$.

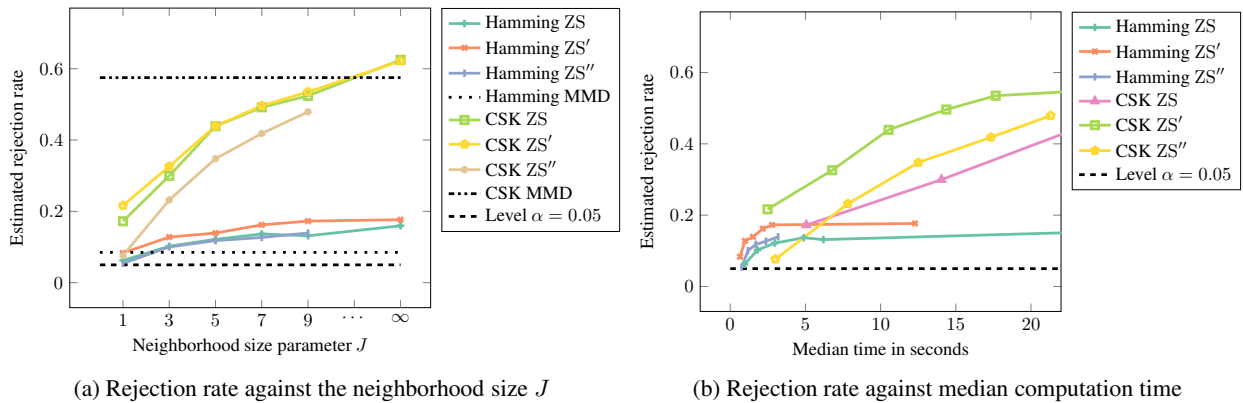
\begin{figure*}
    \centering
    \begin{subfigure}{.48\linewidth}
        \centering
        \begin{tikzpicture}[scale=0.7]
            \begin{axis}
                [
                xlabel=Neighborhood size parameter $J$,
                ylabel=Estimated rejection rate,
                legend pos=outer north east,
                legend cell align=left,
                cycle list name=linecyclelist,
                every axis plot/.append style={
                    line width=1.5pt,
                },
                ymin=0, ymax=0.7,
                xmin=0, xmax=13,
                enlargelimits=0.1,
                ytick={0,0.2,0.4,0.6,0.8},
                xtick={1,3,5,7,9},
                extra x ticks={11,13},
                extra x tick labels={$\cdots$, $\infty$},
                ]
                \addplot coordinates {
                    (1, 0.06229166666666670)
                    (3, 0.10229166666666700)
                    (5, 0.12166666666666700)
                    (7, 0.1366666666666670)
                    (9, 0.13145833333333300)
                    (13, 0.15958333333333300)
                } ;
                \addlegendentry{Hamming $\mathrm{ZS}$};
                \addplot coordinates {
                    (1, 0.08354166666666670)
                    (3, 0.12791666666666700)
                    (5, 0.13916666666666700)
                    (7, 0.16208333333333300)
                    (9, 0.17270833333333300)
                    (13, 0.17667)
                } ;
                \addlegendentry{Hamming $\mathrm{ZS}'$};
                \addplot coordinates {
                    (1, 0.05416666666666670)
                    (3, 0.10020833333333300)
                    (5, 0.11791666666666700)
                    (7, 0.126875)
                    (9, 0.139375)
                } ;
                \addlegendentry{Hamming $\mathrm{ZS}''$};
                \addplot[loosely dotted,no markers] coordinates{ (0, 0.085) (13, 0.085)};
                \addlegendentry{Hamming MMD};
                \addplot coordinates {
                    (1, 0.1722916666666670)
                    (3, 0.29958333333333300)
                    (5, 0.4387500000000000)
                    (7, 0.4916666666666670)
                    (9, 0.5239583333333330)
                    (13, 0.6247916666666670)
                } ;
                \addlegendentry{CSK $\mathrm{ZS}$};
                \addplot coordinates {
                    (1, 0.21645833333333300)
                    (3, 0.3260416666666670)
                    (5, 0.43916666666666700)
                    (7, 0.49625)
                    (9, 0.5350000000000000)
                    (13, 0.62125)
                } ;
                \addlegendentry{CSK $\mathrm{ZS}'$};
                \addplot coordinates {
                    (1, 0.07666666666666670)
                    (3, 0.23187500000000000)
                    (5, 0.3477083333333330)
                    (7, 0.4183333333333330)
                    (9, 0.47916666666666700)
                } ;
                \addlegendentry{CSK $\mathrm{ZS}''$};
                \addplot[dash dot dot,no markers] coordinates{ (0, 0.5749999999999998) (13, 0.5749999999999998)};
                \addlegendentry{CSK MMD};
                \addplot[dashed,no markers] coordinates{ (0, 0.05) (13, 0.05) };
                \addlegendentry{Level $\alpha=0.05$};
            \end{axis}
        \end{tikzpicture}
        \caption{Rejection rate against the neighborhood size $J$}
        \label{fig:simple-neigh-size-zanella-by-size}
    \end{subfigure}
    \begin{subfigure}{.48\linewidth}
        \centering
        \begin{tikzpicture}[scale=0.7]
            \begin{axis}
            [
                xlabel=Median time in seconds,
                ylabel=Estimated rejection rate,
                legend pos=outer north east,
                cycle list name=linecyclelist,
                every axis plot/.append style={
                    line width=1.5pt,
                },
                ymin=0, ymax=0.7,
                legend cell align=left,
                xmax=20,
                enlargelimits=0.1,
                ytick={0,0.2,0.4,0.6,0.8},
            ]
                \addplot coordinates {
                    (0.95471549125     , 0.06229166666666670)
                    (1.81412334375     , 0.10229166666666700)
                    (2.96713807875     , 0.12166666666666700)
                    (4.89214152125     , 0.1366666666666670)
                    (6.19386350375     , 0.13145833333333300)
                    (29.667374402500000, 0.15958333333333300)
                } ;
                \addlegendentry{Hamming $\mathrm{ZS}$};
                \addplot coordinates {
                    (0.6504077725000000, 0.08354166666666670)
                    (0.98039423, 0.12791666666666700)
                    (1.51374236875, 0.13916666666666700)
                    (2.16435998125, 0.16208333333333300)
                    (2.79093509125, 0.17270833333333300)
                    (12.287169676250000, 0.17666666666666700)
                } ;
                \addlegendentry{Hamming $\mathrm{ZS}'$};
                \addplot coordinates {
                    (0.75782358375     , 0.05416666666666670)
                    (1.17815661875     , 0.10020833333333300)
                    (1.7006453987500000, 0.11791666666666700)
                    (2.3772049812500000, 0.126875)
                    (3.1853130225      , 0.139375)
                } ;
                \addlegendentry{Hamming $\mathrm{ZS}''$};
                \addplot coordinates {
                    (5.0570877975000000, 0.1722916666666670)
                    (14.046775021250000, 0.29958333333333300)
                    (22.800015331250000, 0.4387500000000000)
                    (30.836226698750000, 0.4916666666666670)
                    (37.680731532500000, 0.5239583333333330)
                    (121.54186622625   , 0.6247916666666670)
                } ;
                \addlegendentry{CSK $\mathrm{ZS}$};
                \addplot coordinates {
                    (2.477508345, 0.21645833333333300)
                    (6.7807062062500000, 0.3260416666666670)
                    (10.53032518875, 0.43916666666666700)
                    (14.369071126250000, 0.49625)
                    (17.65654460625, 0.5350000000000000)
                    (54.7551323825, 0.62125)
                } ;
                \addlegendentry{CSK $\mathrm{ZS}'$};
                \addplot coordinates {
                    (3.0050541112500000, 0.07666666666666670)
                    (7.819149136250000 , 0.23187500000000000)
                    (12.5193442325     , 0.3477083333333330)
                    (17.348477465000000, 0.4183333333333330)
                    (21.309915431250000, 0.47916666666666700)
                } ;
                \addlegendentry{CSK $\mathrm{ZS}''$};
                \addplot[dashed,no markers] coordinates{ (0, 0.05) (100, 0.05) };
                \addlegendentry{Level $\alpha=0.05$};
            \end{axis}
        \end{tikzpicture}
        \caption{Rejection rate against median computation time}
        \label{fig:simple-neigh-size-zanella-by-time}
    \end{subfigure}
    \caption{%
        Estimated rejection rate of three different families of Zanella-Stein test with varying neighborhood size
        and underlying kernel.
        We evaluate the tests on 12 different synthetic testing scenarios,
        evaluating each test multiple times on independently sampled synthetic datasets.
        To estimate the overall rejection rate,
        we report a simple average across all test evaluations.
        Test power increases as we increase the neighborhood size. 
        We see diminishing returns particularly
        when the Hamming kernel is used to construct the test.
    }
    \label{fig:simple-neigh-size}
\end{figure*}

The structure proposed above is a minimal neighborhood choice that guarantees strong connectivity. 
Indeed, we can edit the length of a sequence, and change the character in each position to a desired one because of the strong connectivity in the intra-length transition.
As we will see in Section~\ref{sec:experiment}, however, this minimal choice may not produce a powerful test,
as it only uses information of neighboring length sequences.
For example, the tail insertion operation by $\mathcal{I}_x$ only encodes the structure of the target distribution
in terms of the end of the sequence.
For sequential models such as  Markov chain models, the longer a sequence is, the more evidence we should have in favor of, or against, a given model.
Editing only the final element of a sequence would therefore not make use of such accumulated evidence.
Indeed, this approach performs relatively weakly in our
benchmark experiments (Figure \ref{fig:simple-neigh-size-zanella-by-time}, case of $J=1$).

\paragraph{Proposed Neighborhood}
Based on the above observation, we generalize the above neighborhood set.
Specifically, we consider expanding the neighborhood set by modifying sequence elements at different locations:
we propose the $J$-location modification neighborhood
\begin{equation}
    \partial_{x_{(1:\ell)}}^J = \cup_{j=1}^J \bigl ( \mathcal{I}_{x_{(1:\ell)}, j}  \cup \mathcal{D}_{x_{(1:\ell)}, j} \cup \mathcal{R}_{x_{(1:\ell)}, j} \bigr ), \label{eqn:J-loc-neigh}
\end{equation}
where
\begin{align*}
    & \mathcal{I}_{x_{(1:\ell)}, j} \subset \alphabet^{(\ell+1)} = \{\insAction(x_{(1:\ell)},j,s): s  \in {\mathcal{N}}_{\mathrm{\insAction}} \},\\
    & \mathcal{D}_{x_{(1:\ell)}, j} \subset \alphabet^{(\ell-1)} = \{\delAction(x_{(1:\ell)},j) \text{ if } x_{\ell-j+1}\in \mathcal{N}_{\insAction}\},\\
    & \mathcal{R}_{x_{(1:\ell)}, j} \subset \alphabet^{(\ell)} = \{\repAction(x_{(1:\ell)},j,s):  s\in \mathcal{N}_{\repAction}(x_\ell) \}.
\end{align*}
Here, $\insAction(x_{(1:\ell)}, j, s)$ denotes the sequence extending $x_{(1:\ell)}$ by inserting $s$ at the $(j-1)$-th location counting back from the end of the string, using symbols from a fixed set $\mathcal{N}_{\insAction}\subset\alphabet$;
    $\delAction(x_{(1:\ell)},j)$ deletes the $(j-1)$-th element counting back from the end of $x_{(1:\ell)}$; and
    $\repAction(x_{(1:\ell)},j,s)$  replaces by $s$ the $(j-1)$-th element from the end of $x_{(1:\ell)}$.
    Note that the deletion and insertion operations must be paired to ensure the symmetry of the graph.
As with the substitution set $\mathcal{N}_{\repAction}$, we may choose $\mathcal{N}_{\insAction}$ such that it depends on the point $x$;
the resulting neighborhood requires additional care to maintain symmetry and (optionally) strong connectivity. See \cref{sec:additional-exp-conditional} for an example.

\paragraph{Test Procedure} Using one of the proposed neighborhoods for the Zanella-Stein operator, we can define a KSD goodness-of-fit test for sequential models. 
Given an i.i.d. data $\{X_i\}_{i=1}^n \iidsim \dis$, we consider testing the null $H_0: \ksdmod{\dis}{\dit}=0$ against the alternative $H_1: \ksdmod{\dis}{\dit}\neq 0$.
Note that when the KSD distinguishes any distributions, the null and alternative become $H_0: \dit=\dis$ and $H_1:\dit\neq\dis$, respectively.
Below, we present two tests, which differ in how they compute test thresholds.

The first test is the parametric bootstrap test in Algorithm~\ref{alg:param-bootstrap}.
The parametric bootstrap repeatedly draws samples from the model $\dit$ and simulates the distribution of the statistic \eqref{eqn:ustat}.
Because of this feature, this procedure does not apply if sampling from the model is infeasible.

The second test is the wild bootstrap test described in Algorithm~\ref{alg:wild-bootstrap}, which is analogous to the existing KSD tests~\citep{chwialkowski_kernel_2016,liu_kernelized_2016,yang_goodness--fit_2018}.
As we have seen in Section~\ref{sec:background}, the asymptotic distribution of the (scaled) KSD estimate \eqref{eqn:ustat} is known.
The wild bootstrap test simulates this distribution by repeatedly computing the following quantity:%
\begin{equation}
    \ksdwildustat{n}{\dit}{ D_n }{ W  }\coloneqq \frac{\sum_{i \neq j}^n (W_i-1) (W_j-1) h_{\pmft} (X_i, X_j)}{ n(n-1) }\ 
\end{equation}
where $W=(W_1, \dots, W_n)$ is a sample from a multinomial distribution $\mathrm{Multi}(n; 1/n,\dots,1/n)$ with $n$ trials and $n$ events and independent of the data $D_n=\{X_1,\dots,X_n\}$.
The test has an only asymptotically correct size~\citep[][Theorem 3.1]{dehling_random_1994}, and may not be suitable for a small sample size $n$.
In contrast to the parametric bootstrap test, this test does not require sampling from the model, and only needs single evaluation of the Stein kernel $h_p$ over the data. 
Treating the evaluation of the Stein kernel as constant, the test's computational complexity is $O(n^2)$, since the computation of the KSD statistic is 
$O(n^2)$; this computation can be paralellized.

    \section{Experiments}\label{sec:experiment}

    We investigate the performance of the proposed test through synthetic experiments with defined ground truths.\footnote{The code is available at \url{https://github.com/test-for-sequential-models/code}}
In particular, we aim to answer the following questions:
(a)~how leveraging model structure using a Stein kernel improves test performance over 
the maximum mean discrepancy (MMD)~\citep[][a purely sample-based kernel discrepancy where no model information is used]{gretton_kernel_2012};
(b)~the effect of neighborhood choice on the KSD test power.
In the following, we use the Barker
balancing function $\balfun{t}=t/(1+t)$ to define the KSD.

Throughout our experiments, we will make use
of two simple kernels.
The first is the exponentiated Hamming kernel,
$k_{\mathrm{H}}(x_{(1:\ell)},y_{(1:\ell)})\coloneqq \exp\{-{\ell}^{-1} \sum_{i=1}^\ell \delta_{x_i}\of{y_i}\}$,
where $\delta_{x}\of{y}=1$ if $x=y$ and $0$ otherwise.
Additionally, we define
$k_{\mathrm{H}}(x_{(1:\ell)},y_{(1:\ell')})=0$
whenever $\ell\ne\ell'$.
The second is a string kernel which we refer to as the
\emph{contiguous subsequence kernel} (CSK).
This kernel counts the number of contiguous subsequences
of a particular length, present in the two input sequences,
\begin{align*}
    k_{\mathrm{CSK}}\of{x, y} &\coloneqq
    \frac{k_\text{u}(x,y)}{\sqrt{%
        k_\text{u}(x,x)
        \cdot
        k_\text{u}(y,y)
    }},
\end{align*}
where
\begin{align*}
    k_\text{u}(x_{(1:\ell)},y_{(1:\ell')}) &\coloneqq
    \sum_{i=1}^{\ell-t+1}
    \sum_{j=1}^{\ell'-t+1}
    \delta_{x_{(i:i+t-1)}}\of{y_{(j:j+t-1)}},
\end{align*}
with parameter $t$ for the subsequence length.
We use the normalized version so that the kernel does not overly depend
on the sequence lengths.
The Hamming kernel does not take into account the sequence structure and can be considered as a naive choice,
whereas the CSK kernel does.

\begin{figure}[t]
    \centering
    \begin{tikzpicture}[scale=0.7]
        \begin{axis}
            [
            xlabel={Holding probability in the ground truth, $p_\text{hold}$},
            ylabel=Estimated rejection rate,
            ymin=0, ymax=1,
            ytick={0,0.2,0.4,0.6,0.8,1},
            xtick={0,0.1,0.2},
            enlargelimits=0.1,
            legend pos=outer north east,
            legend cell align=left,
            cycle list name=linecyclelist,
            every axis plot/.append style={
                line width=1.5pt,
            },
            ]
            \addplot table[
            col sep=comma,
            x={perturbation},
            y={Stein(k=CSK: op=__mathrm{ZS}_{infty:b}'_)},
            ]{figures/exp_random_walk_with_memory_rejection_rate_by_perturbation.csv};
            \addlegendentry{KSD};
            \addplot table[
            col sep=comma,
            x={perturbation},
            y={MMD(k=CSK: n=100)},
            ]{figures/exp_random_walk_with_memory_rejection_rate_by_perturbation.csv};
            \addlegendentry{MMD};
            \addplot table[
            col sep=comma,
            x={perturbation},
            y={LR(oracle)},
            ]{figures/exp_random_walk_with_memory_rejection_rate_by_perturbation.csv};
            \addlegendentry{LR oracle};
            \addplot table[
            col sep=comma,
            x={perturbation},
            y={LR(_k=k__mathrm{model}_)},
            ]{figures/exp_random_walk_with_memory_rejection_rate_by_perturbation.csv};
            \addlegendentry{LR MC};
            \addplot[dashed,no markers] coordinates{ (0, 0.05) (0.25, 0.05) };
            \addlegendentry{Level $\alpha=0.05$};
        \end{axis}
    \end{tikzpicture}
    \caption{%
        Estimated rejection rate (test power)
        of the proposed test
        when the model distribution
        and ground truth differ by a perturbation
        that is unlikely to be observed in any given sample.
    }
    \label{fig:exp-ksd-is-good}
\end{figure}
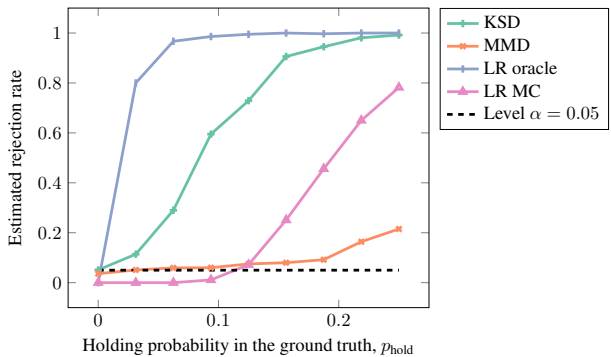

\subsection{Demonstration of Test Power}

We first show that the KSD indeed benefits from information supplied by the model.
For this purpose, we construct a problem where the alternative is given by
adding a slight perturbation to the model.
We choose $\alphabet=\{0,\ldots,m-1\}$
for our alphabet, with $m=8$.
The model distribution
is a random walk through $\alphabet$,
starting from a random uniform element of $\alphabet$,
with increments drawn uniformly from $\{-1,+1\}$.
We use a cyclic structure, where we identify $0$ with $m$
such that the random walk wraps around.
The chain terminates with probability $1/20$ after each step.
We construct an alternative distribution as follows:
At every step, if and only if the current element is $0\in\alphabet$,
we randomly replace the increment with $0$ with a probability
$p_\text{hold}\in[0,1/4]$.
That is, we introduce a random holding step that occurs
on average in $1/(4m)=1/32$ of steps,
when the perturbation parameter is at its maximum.
The perturbation is unlikely to be observed in any given sample,
making the problem challenging.

For the KSD test we use the $J$-location modification neighborhood
\eqref{eqn:J-loc-neigh} 
with $J=\infty$; i.e., we allow edits anywhere in the sequence.
As baselines we use an MMD test that draws 100
samples from the model,
a likelihood-ratio test where the alternative
consists of all first-order Markov chains on $\alphabet$,
and an oracle likelihood-ratio test where the alternative
is the (generally unknown) ground truth distribution.
Both kernel tests are built using the CSK kernel.
We use parametric bootstrap for all four tests.
Tests are evaluated on $n=30$ i.i.d.\ sample points from the perturbed distribution.

In \cref{fig:exp-ksd-is-good} we can see that the KSD
test outperforms the two non-oracle baselines.
At low degrees of perturbation, the problem is challenging
because perturbations are rarely observed.
The KSD test is able to overcome this difficulty
by exploiting information provided by the model.

\subsection{Choice of Neighborhood Graph}\label{sec:choice-of-neighbourhood-graph}

Our test in the preceding experiment
performed well, in part due to a good choice of neighborhood.
We now further investigate the choice of neighborhood
for the Zanella-Stein
operator.
In order to compare the performance
of different choices of Stein kernel,
we constructed 12 synthetic
testing scenarios with various properties, ensuring
our conclusions are broadly applicable, and not specific to properties
of a particular model.
The experiments in this section
will evaluate the average performance
of each test across these scenarios.
We use discrete-time Markov chains for the model $\dit$
and ground truth $\dis$.
The testing scenarios
are fully specified in Appendix~\ref{sec:testing-scenarios} (see \cref{sec:scenarios-table} for individual results). 

More densely connected neighborhood graphs
should yield more powerful tests,
but such tests are slower to evaluate.
To understand the tradeoff between compute budget
and test power,
we construct a family of KSD tests
parameterized by the size of the neighborhood
of each point.
We consider three classes of neighborhood:
(a)~the $J$-location modification neighborhood
$\partial^J_{x_{(1:\ell)}}$, 
(b)~the substitution neighborhood
$\cup_{j\leq J } \mathcal{R}_{x_{(1:\ell)},j}$, 
and (c)~the neighborhood 
$\cup_{j\leq J}
(\mathcal{I}_{x_{(1:\ell)},j} \cup
\mathcal{D}_{x_{(1:\ell)},j})$ of insertions and deletions.
In each case, the symbol neighborhoods $\mathcal{N}_\text{ins}$
and $\mathcal{N}_{\repAction}$ are set to the entire alphabet $\alphabet$.
We denote the resulting Stein operators by $\mathrm{ZS}$,
$\mathrm{ZS}'$, and $\mathrm{ZS}''$, respectively.
We construct KSDs by
applying these operators on the two kernels specified earlier. 
To estimate the critical value, we use parametric bootstrap.
We also include a pair of MMD tests, one for each kernel, as a baseline.

Results are in \Cref{fig:simple-neigh-size-zanella-by-size}.
Larger neighborhoods lead to more powerful tests, as expected.
However, \cref{fig:simple-neigh-size-zanella-by-time}
shows that we face diminishing returns when trading off
compute budget against test power.
The CSK kernel always does better than the  Hamming kernel.
Interestingly, for the CSK kernel,
the simpler $\mathrm{ZS}'$ test performs  slightly better than the ZS test,
since in this instance, the larger neighbourhood of the latter
does not yield significant additional information, while entailing a small increase in variance
of the test statistic.
Both of these Stein tests outperform the CSK MMD for $J=\infty$.

\begin{figure}[h]
    \centering
    \begin{tikzpicture}[scale=0.7]
        \begin{axis}
            [
            xlabel=Symbol neighborhood size parameter $d$,
            ylabel=Estimated rejection rate,
            legend pos=outer north east,
            legend cell align=left,
            cycle list name=linecyclelist,
            every axis plot/.append style={
                line width=1.5pt,
            },
            ymin=0, ymax=0.7,
            xmin=1, xmax=5,
            enlargelimits=0.1,
            ytick={0,0.2,0.4,0.6,0.8},
            xtick={1,2,3},
            extra x ticks={4,5},
            extra x tick labels={$\cdots$, $\infty$},
            ]
            \addplot coordinates {
                (1, 0.08354166666666670)
                (2, 0.09000000000000000)
                (3, 0.09041666666666670)
                (5, 0.08354166666666670)
            } ;
            \addlegendentry{Ham. $\mathrm{ZS}', J=1$};
            \addplot coordinates {
                (1, 0.12979166666666700)
                (2, 0.130625)
                (3, 0.12312500000000000)
                (5, 0.12791666666666700)
            } ;
            \addlegendentry{Ham. $\mathrm{ZS}', J=3$};
            \addplot coordinates {
                (1, 0.22270833333333300)
                (2, 0.2160416666666670)
                (3, 0.20833333333333300)
                (5, 0.21645833333333300)
            } ;
            \addlegendentry{CSK $\mathrm{ZS}', J=1$};
            \addplot coordinates {
                (1, 0.32937500000000000)
                (2, 0.34666666666666700)
                (3, 0.3472916666666670)
                (5, 0.3260416666666670)
            } ;
            \addlegendentry{CSK $\mathrm{ZS}', J=3$};
            \addplot[dashed,no markers] coordinates{ (0, 0.05) (13, 0.05) };
            \addlegendentry{Level $\alpha=0.05$};
        \end{axis}
    \end{tikzpicture}
    \caption{%
        Estimated rejection rate of several Zanella-Stein tests based on
        the substitution neighborhood.
        We control the size of the symbol neighborhood $\mathcal{N}_\text{rep}$
        and estimate rejection rate as in \cref{fig:simple-neigh-size}.
    }
    \label{fig:zanella-growing-alphabet}
\end{figure}
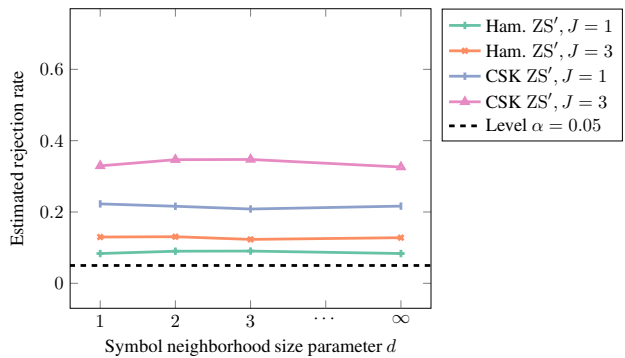

Another approach to controlling the size and connectivity
of the neighborhood is to control the symbol neighborhoods
$\mathcal{N}_\text{ins}$ and $\mathcal{N}_\text{rep}$.
We consider the operator $\mathrm{ZS}'$ (a neighborhood consisting of substitutions) 
with a cyclic structure over the alphabet $\alphabet$ 
and the symbol neighborhood $\mathcal{N}_\text{rep}\of{s}$
limited by the distance in the cyclic alphabet.
In \cref{fig:zanella-growing-alphabet} we can see no notable impact
of symbol neighborhood on test power.
This implies that we may construct tests using a sparse neighborhood.
The experiment also shows that changes to the size of the neighborhood
are not sufficient to fully explain our previous results.
Test power is evidently also impacted by the location
of edits rather than just the number.

\subsection{Use of Evidence in Long Sequences}

Edits across the entire sequence lead to higher test power.
One explanation for this effect is that editing the entire sequence,
instead of just editing near the end,
allows the test to make better use of the evidence available from longer sequences.
Where the data are distributed according to a Markov chain or some other model that mixes quickly,
longer sequences provide  more information about the distribution.
We now show that the proposed test can indeed make use of such information.

In order to illustrate this feature, we construct the following problem:
The model distribution and ground truth are second-order Markov chains
over a finite alphabet with $|\alphabet|=8$.
We select two transition kernels randomly from a Dirichlet distribution
with concentration parameter $\alpha=1$.
The model distribution uses one of these two kernels, 
whereas the ground truth uses an even mixture of the two kernels,
so that at each step each of the kernels is used with equal probability.
The two chains both stop with a fixed probability $1/\lambda$ after each step.
We fix the sample size $n$ at 8
and vary the stopping probability in order to control the length of the samples.

\begin{figure}[t!]
    \centering
    \begin{tikzpicture}[scale=0.7]
        \begin{axis}
            [
            xlabel={Expected sequence length $\lambda$}, %
            ylabel=Estimated rejection rate,
            ymin=0, ymax=1,
            ytick={0,0.2,0.4,0.6,0.8,1},
            enlargelimits=0.1,
            legend pos=outer north east,
            legend cell align=left,
            cycle list name=linecyclelist,
            every axis plot/.append style={
                line width=1.5pt,
            },
            ]
            \addplot table[
            col sep=comma,
            x={E_length},
            y={Stein(k=CSK: op=__mathrm{ZS}_{infty:b}'_)},
            ]{figures/exp_snd_order_mc_few_long_samples_rejection_rate_by_E_length.csv};
            \addlegendentry{KSD};
            \addplot table[
            col sep=comma,
            x={E_length},
            y={MMD(k=CSK: n=100)},
            ]{figures/exp_snd_order_mc_few_long_samples_rejection_rate_by_E_length.csv};
            \addlegendentry{MMD};
            \addplot table[
            col sep=comma,
            x={E_length},
            y={LR(oracle)},
            ]{figures/exp_snd_order_mc_few_long_samples_rejection_rate_by_E_length.csv};
            \addlegendentry{LR oracle};
            \addplot table[
            col sep=comma,
            x={E_length},
            y={LR(_k=k__mathrm{model}_)},
            ]{figures/exp_snd_order_mc_few_long_samples_rejection_rate_by_E_length.csv};
            \addlegendentry{LR MC};
            \addplot[dashed,no markers] coordinates{ (2, 0.05) (20, 0.05) };
            \addlegendentry{Level $\alpha=0.05$};
        \end{axis}
    \end{tikzpicture}
    \caption{%
        Estimated rejection rate of the proposed test, 
        where we control the expected length of the samples,
        while holding the sample count fixed at 8.
        As the expected length increases, test power grows.
    }
    \label{fig:exp-few-long-sequences}
\end{figure}
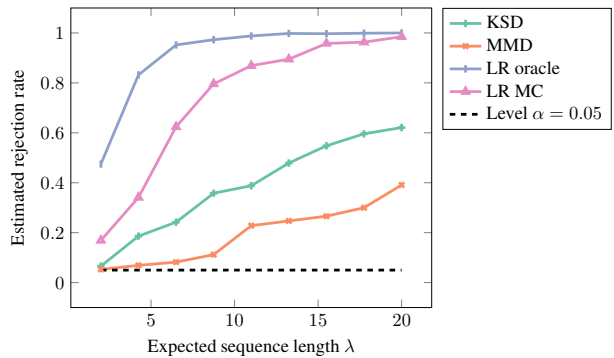

The KSD test for this experiment is constructed using a large, strongly connected 
neighborhood: specifically, the $J$-location modification neighborhood with $J=\infty$,
which allows insertion, deletion, and substitution anywhere in the sequence.
As a baseline, we compare against an MMD test using
100 samples from the model distribution,
as well as two likelihood-ratio tests:
an oracle test, %
and a test where the alternative
consists of all second-order Markov chains over $\alphabet$.
The two kernel tests are constructed using the CSK kernel
with subsequence length $t=3$,
and we use parametric bootstrap for all tests.

The results are shown in \cref{fig:exp-few-long-sequences}.
We can see that the test is able to reject more often
on longer sequences, without the need for more samples, and outperforms the MMD test.

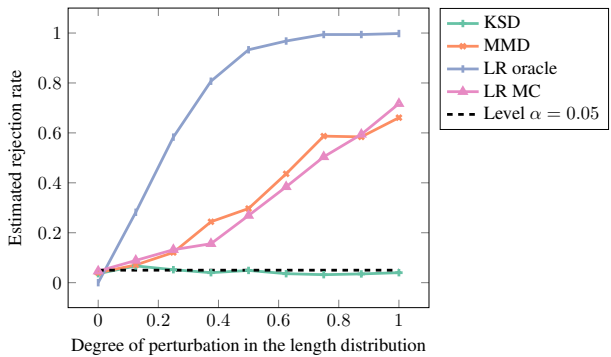
\begin{figure}[t]
    \centering
    \begin{tikzpicture}[scale=0.7]
        \begin{axis}
            [
            xlabel=Degree of perturbation in the length distribution, %
            ylabel=Estimated rejection rate,
            ymin=0, ymax=1,
            ytick={0,0.2,0.4,0.6,0.8,1},
            enlargelimits=0.1,
            legend pos=outer north east,
            legend cell align=left,
            cycle list name=linecyclelist,
            every axis plot/.append style={
                line width=1.5pt,
            },
            ]
            \addplot table[
            col sep=comma,
            x={perturbation},
            y={Stein(k=CSK: op=__mathrm{ZS}_{infty:b}_)},
            ]{figures/exp_random_mc_varied_length_dist_rejection_rate_by_perturbation.csv};
            \addlegendentry{KSD};
            \addplot table[
            col sep=comma,
            x={perturbation},
            y={MMD(k=CSK: n=100)},
            ]{figures/exp_random_mc_varied_length_dist_rejection_rate_by_perturbation.csv};
            \addlegendentry{MMD};
            \addplot table[
            col sep=comma,
            x={perturbation},
            y={LR(oracle)},
            ]{figures/exp_random_mc_varied_length_dist_rejection_rate_by_perturbation.csv};
            \addlegendentry{LR oracle};
            \addplot table[
            col sep=comma,
            x={perturbation},
            y={LR(_k=k__mathrm{model}_)},
            ]{figures/exp_random_mc_varied_length_dist_rejection_rate_by_perturbation.csv};
            \addlegendentry{LR MC};
            \addplot[dashed,no markers] coordinates{ (0, 0.05) (1, 0.05) };
            \addlegendentry{Level $\alpha=0.05$};
        \end{axis}
    \end{tikzpicture}
    \caption{
        Estimated rejection rate of the proposed KSD test.
        We use a randomly generated Markov chain as the model distribution
        and perturb the distribution of lengths for the data. %
        The proposed KSD test fails to reject in this setting.
    }
    \label{fig:exp-length-distribution}
\end{figure}

\subsection{Sensitivity to Changes in Length}

There is a caveat with the family of neighborhoods we have described:
these neighborhoods are not particularly sensitive
to changes in the distribution %
over sequence lengths. We demonstrate this point by choosing a randomly generated
Markov chain over a finite alphabet $|\alphabet|=10$
as the model distribution, with the transition kernel again chosen from a Dirichlet
distribution as above.
Here, the ground truth and model use the same transition kernel,
but we vary the stopping probability.
The stopping probability is selected such that the expected
sequence length is 8 under the model distribution, 
whereas we vary the sequence length between 8 and 20 under the ground truth. 

We use the same tests as in the previous experiment,
except that we use a subsequence length $t$ of 2 for the CSK kernel
and the alternative for the LR baseline now consists
of first- rather than second-order Markov chains. 

In \cref{fig:exp-length-distribution},
we can see
that the KSD test fails to reject entirely, outperformed by both types of baseline.
This experimental result highlights the need for bespoke
tests,
when we desire that a test is sensitive to a specific aspect
of the distribution.
A test user who is particularly interested in the length distribution
would likely benefit from directly testing
the distribution over lengths using a classical test.

\subsection{Testing Intractable Models}\label{sec:mrf-experiment}

A major benefit of the proposed test is that it does not require us
to sample from the model distribution, nor to compute a normalized density.
Here, we showcase our test using a Markov Random Field (MRF) model, a class of intractable models. 
An MRF model $P$ is defined by normalizing an exponentiated \emph{potential function} $h$, 
i.e., we have $\pmft(x) \propto \exp(h(x))$. 
The normalization constant is often intractable, and hence so is the probability mass function $\pmft$. 

We design an MRF model by specifying a single potential $h$ across the entire
sequence space $\samplespace$:
\begin{align*}
    h\of{x_{(1:\ell)}} &\coloneqq
    \begin{cases}
        C\cdot \ell + \theta \sum_{i=1}^{\ell-1} \delta_{x_i}\of{x_{i+1}}, & \text{if } \ell \le M \\
        -\infty, & \text{else.}
    \end{cases}
\end{align*}
In our experiment below, we hold the length parameters $C$ and $M$ fixed, while perturbing
the distribution by varying $\theta$.
The parameter $\theta$ is a concentration parameter
controlling the correlation between successive elements of a sequence; 
symbol repetitions become more common as $\theta$ grows. 

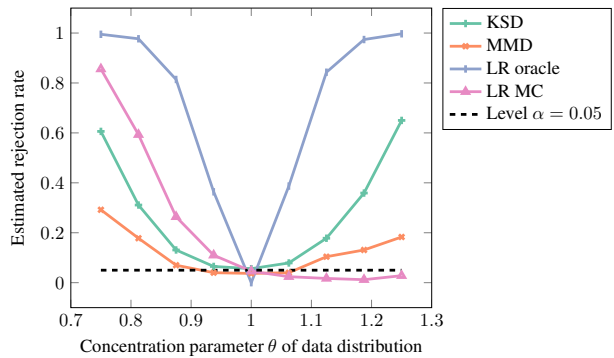
\begin{figure}[t]
    \centering
    \begin{tikzpicture}[scale=0.7]
        \begin{axis}
            [
            xlabel=Concentration parameter $\theta$ of data distribution,
            ylabel=Estimated rejection rate,
            ymin=0, ymax=1,
            ytick={0,0.2,0.4,0.6,0.8,1},
            enlargelimits=0.1,
            legend pos=outer north east,
            legend cell align=left,
            cycle list name=linecyclelist,
            every axis plot/.append style={
                line width=1.5pt,
            },
            ]
            \addplot table[
            col sep=comma,
            x={sample_concentration_param},
            y={Stein-Wild(k=CSK: op=__mathrm{ZS}_{infty:b}_)},
            ]{figures/exp_simple_mrf_rejection_rate_by_sample_concentration_param.csv};
            \addlegendentry{KSD};
            \addplot table[
            col sep=comma,
            x={sample_concentration_param},
            y={MMD-Wild(k=CSK: n=same)},
            ]{figures/exp_simple_mrf_rejection_rate_by_sample_concentration_param.csv};
            \addlegendentry{MMD};
            \addplot table[
            col sep=comma,
            x={sample_concentration_param},
            y={LR(oracle)},
            ]{figures/exp_simple_mrf_rejection_rate_by_sample_concentration_param.csv};
            \addlegendentry{LR oracle};
            \addplot table[
            col sep=comma,
            x={sample_concentration_param},
            y={LR(_k=k__mathrm{model}_)},
            ]{figures/exp_simple_mrf_rejection_rate_by_sample_concentration_param.csv};
            \addlegendentry{LR MC};
            \addplot[dashed,no markers] coordinates{ (0.75, 0.05) (1.25, 0.05) };
            \addlegendentry{Level $\alpha=0.05$};
        \end{axis}
    \end{tikzpicture}
    \caption{
        Performance comparison 
        using a simple MRF model with a concentration parameter $\theta$ varied; 
        the parameter $\theta$ is fixed at $\theta=1$ for the model, and 
        is chosen from $0.75$ to $1.25$ for the sample distribution. 
        Our proposed test outperforms both non-oracle baselines in one region and
        outperforms MMD in the other region.
    }
    \label{fig:exp-simple-mrf}
\end{figure}

Our problem is given as follows. 
We specify $\theta=1$ for the model distribution and vary $\theta \in[0.75,1.25]$
for the data distribution; the setup tests the sensitivity to this perturbation. 
We specify $M=20$ in all cases,
and vary $C$ with $\theta $ so that the mean length is fixed at 10
and the length distribution does not depend on $\theta$.
The alphabet size is 3.
We compare the proposed test
against the MMD and LR baselines.
Note that although it is generally challenging to sample from MRF models 
or to compute their normalized densities, 
the above special model allows us to perform both operations, and hence MMD and LR tests. 
For the LR baseline we use an oracle test (with access to the data distribution),
as well as a test with $H_1$ consisting of first-order Markov chains; 
the latter baseline does not contain the above MRF family in the alternative.
The MMD and KSD tests use the CSK kernel
with subsequence length $t=2$.

The proposed test performs competitively with the non-oracle baselines.
In \cref{fig:exp-simple-mrf}
we can see that the proposed test outperforms the MMD baseline
across the full range of perturbations. 
In particular, it outperforms the non-oracle LR baseline in one region; 
the LR baseline fails as $\theta$ grows, due to lack of consistency.
We emphasize that our proposed test applies more generally
than the LR test, since we do not require normalized densities.
Moreover, using the wild bootstrap procedure,
we do not need to sample from the model, unlike the MMD test.

\begin{figure}[t]
    \centering
    \begin{tikzpicture}[scale=0.7]
        \begin{axis}
        [
            xlabel=Number of samples,
            ylabel=Estimated rejection rate,
            ymin=0, ymax=1,
            ytick={0,0.2,0.4,0.6,0.8,1},
            enlargelimits=0.1,
            legend pos=outer north east,
            legend cell align=left,
            cycle list name=linecyclelist,
            every axis plot/.append style={
                line width=1.5pt,
            },
        ]
            \addplot table[
            col sep=comma,
            x={N_sequences},
            y={Stein-Wild(k=CSK: op=__mathrm{ZS}_{infty:b}_)},
            ]{figures/exp_simple_mrf_rejection_rate_by_N_sequences.csv};
            \addlegendentry{KSD};
            \addplot table[
            col sep=comma,
            x={N_sequences},
            y={MMD-Wild(k=CSK: n=same)},
            ]{figures/exp_simple_mrf_rejection_rate_by_N_sequences.csv};
            \addlegendentry{MMD};
            \addplot table[
            col sep=comma,
            x={N_sequences},
            y={LR(oracle)},
            ]{figures/exp_simple_mrf_rejection_rate_by_N_sequences.csv};
            \addlegendentry{LR oracle};
            \addplot table[
            col sep=comma,
            x={N_sequences},
            y={LR(_k=k__mathrm{model}_)},
            ]{figures/exp_simple_mrf_rejection_rate_by_N_sequences.csv};
            \addlegendentry{LR MC};
            \addplot[dashed,no markers] coordinates{ (200, 0.05) (700, 0.05) };
            \addlegendentry{Level $\alpha=0.05$};
        \end{axis}
    \end{tikzpicture}
    \caption{
        Performance of the proposed KSD test
        on a simple MRF model.
        We perform the same experiment as shown in \cref{fig:exp-simple-mrf},
        in this case holding fixed $\theta=0.9$ and varying the number of samples.
        In this region, our proposed test outperforms the MMD baseline but not the LR baseline.
    }
    \label{fig:exp-simple-mrf-power-curve}
\end{figure}
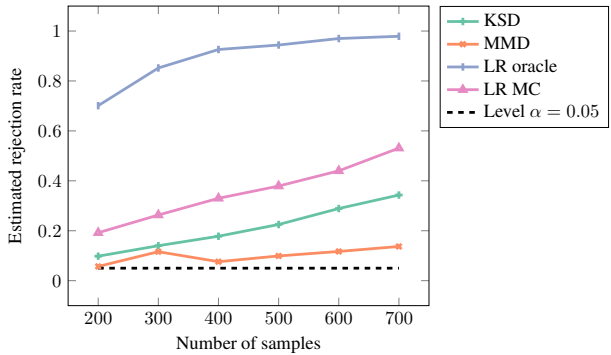

We also show how test power improves with the data size 
in \cref{fig:exp-simple-mrf-power-curve}.
We choose the region where our test outperforms the MMD baseline, but not the LR baseline.
The parameter $\theta$ is fixed to $0.9$, while we vary the sample size $n$. 
We can see that the MMD baseline has a low sample efficiency for this problem, confirming again 
the MMD's inability to use the model structure. 
Both the LR baseline and our proposed KSD test make use of model information.
Only the KSD test can work in the absence of a normalized density, however.

\subsection{Inspecting Output-Constrained Generative Models}\label{sec:additional-exp-conditional}
In practical applications, we are often interested in the ability of a model to output a particular set of patterns or values.
For example, in language modeling, one might be interested in the ability to generate a subset of a language or sentences of a particular form (e.g., questions).
Mathematically, we can state this objective as follows:
for a model $\dit$, we aim to evaluate the goodness-of-fit of its conditional distribution $P_{A},$
derived by restricting the model to a set $A\subset \samplespace$ (see below for an example of $A$).
The probability mass function (pmf) $\pmft_A$ of the conditional $\dit_A$ is given by
\[ \pmft_A (x) = \frac{\pmft(x)}{\sum_{\tilde{x} \in A} \pmft(\tilde{x})},\]
where $\pmft$ is the pmf of $\dit$.
The normalization constant appearing in $\pmft_A$ is typically unknown (even when $p$ is normalized)
and thus as in the MRF experiment, this task is challenging both for MMD and likelihood-based diagnostics.

\begin{figure}[t]
    \centering
    \begin{tikzpicture}[scale=0.7]
        \begin{axis}
            [
            xlabel={Mix-parameter $\pi$},
            ylabel=Estimated rejection rate,
            ymin=0, ymax=1,
            ytick={0,0.2,0.4,0.6,0.8,1},
            xmin=0, xmax=0.5,
            enlargelimits=0.1,
            legend pos=outer north east,
            legend cell align=left,
            cycle list name=linecyclelist,
            every axis plot/.append style={
                line width=1.5pt,
            },
            ]
            \addplot table[
            col sep=comma,
            x={perturbation_sample_size_50},
            y={MMD-Wild(n=100)},
            ]{figures/exp_language_rejection_rate_by_perturbation_sample_size_50.csv};
            \addlegendentry{MMD};
            \addplot table[
            col sep=comma,
            x={perturbation_sample_size_50},
            y={Stein-Wild(sub_ins_and_del')},
            ]{figures/exp_language_rejection_rate_by_perturbation_sample_size_50.csv};
            \addlegendentry{CSK $\mathrm{ZS}$};
            \addplot table[
            col sep=comma,
            x={perturbation_sample_size_50},
            y={Stein-Wild(sub_only)},
            ]{figures/exp_language_rejection_rate_by_perturbation_sample_size_50.csv};
            \addlegendentry{CSK $\mathrm{ZS}'$};
            \addplot[dashed,no markers] coordinates{ (0, 0.05) (0.5, 0.05) };
            \addlegendentry{Level $\alpha=0.05$};
        \end{axis}
    \end{tikzpicture}
    \caption{%
        Performance comparison in the language model experiment. 
        The two KSD tests outperform the MMD test. 
        In particular, the $\mathrm{ZS}$ variant beats the $\mathrm{ZS}'$ one when the mixture proportion $\pi$ is small, 
        showing a benefit of the richer neighborhood structure. 
    }
    \label{fig:exp-lanuage}
\end{figure}
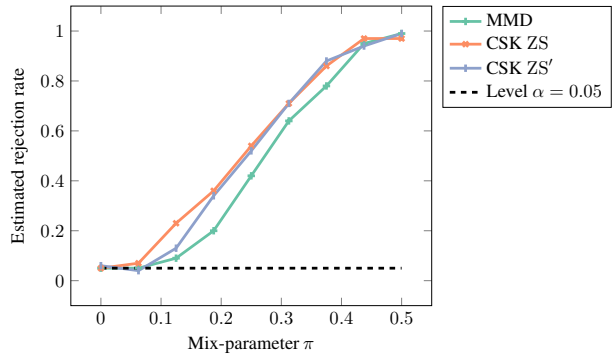

Following the above setup, we conduct an additional experiment.
Specifically, we consider evaluating the ability of a language model
to generate \emph{question sentences} (i.e.,
$A$ is the set of sequences ending with symbol '?').
Our task is detecting the departure of one model configuration from its mixture with another (with the mixture parameter $\pi$ varied).
We use the pretrained model \texttt{gpt2}~\citep{radford2019language} to obtain two configurations: these are attained by inputting two different prompts:
\texttt{How are} and \texttt{Where is}.
To sample from the language model,
we generate a sequence of up to 5 additional tokens following the prompt,
stopping when we encounter the token '?'.
The sample is rejected if this token is not encountered within 5 steps following
the prompt.
For each prompt, this gives a different distribution over question sentences.
The model distribution is specified as the output with the prompt \texttt{How are},
while the ground truth is a mixture of outputs by randomly selecting between the two prompts.
We draw $n=50$ samples. 

We compare the KSD against the MMD test.
For the MMD, we generate samples using rejection sampling.
This is tractable in this experiment, because the prompts are designed
to yield a low rejection probability (questions are likely to be generated).
For the KSD, we use the $\mathrm{ZS}$ and $\mathrm{ZS}'$ configurations as in \cref{sec:choice-of-neighbourhood-graph} with $J=1$. 
We choose sparse, point-dependent insertion and substitution neighborhoods; we use 16 high-probability tokens for a given context,
and thus do not use the sparsification approach from \cref{sec:choice-of-neighbourhood-graph}.
Both MMD and KSD tests use wild bootstrap procedures.
Here, we do not include likelihood-based tests
as they are intractable in this setting.
For further experimental details, see \Cref{sec:toy-experiment-details}.
\Cref{fig:exp-lanuage} shows the result, showcasing the KSD's utility in a challenging task.

\section*{Acknowledgements}
We thank the three anonymous referees for their helpful feedback. 
HK and AG acknowledge the financial support by the Gatsby Charitable Foundation.

\bibliography{./icml.bib}

\begin{thebibliography}{39}
\providecommand{\natexlab}[1]{#1}
\providecommand{\url}[1]{\texttt{#1}}
\expandafter\ifx\csname urlstyle\endcsname\relax
  \providecommand{\doi}[1]{doi: #1}\else
  \providecommand{\doi}{doi: \begingroup \urlstyle{rm}\Url}\fi

\bibitem[Anastasiou et~al.(2021)Anastasiou, Barp, Briol, Ebner, Gaunt,
  Ghaderinezhad, Gorham, Gretton, Ley, Liu, Mackey, Oates, Reinert, and
  Swan]{anastasiou_steins_2021}
Anastasiou, A., Barp, A., Briol, F.-X., Ebner, B., Gaunt, R.~E., Ghaderinezhad,
  F., Gorham, J., Gretton, A., Ley, C., Liu, Q., Mackey, L., Oates, C.~J.,
  Reinert, G., and Swan, Y.
\newblock Stein's method meets computational statistics: A review of some
  recent developments.
\newblock \emph{Statistical Science}, 38\penalty0 (1), 2021.
\newblock \doi{10.1214/22-sts863}.

\bibitem[Aronszajn(1950)]{aronszajn_theory_1950}
Aronszajn, N.
\newblock Theory of reproducing kernels.
\newblock \emph{Transactions of the American mathematical society}, 68\penalty0
  (3):\penalty0 337--404, 1950.

\bibitem[Barbour(1988)]{barbour_steins_1988}
Barbour, A.~D.
\newblock {S}tein's method and {Poisson} process convergence.
\newblock \emph{Journal of Applied Probability}, 25:\penalty0 175--184, 1988.
\newblock Publisher: Cambridge University Press.

\bibitem[Barker(1965)]{Barker_1965}
Barker, A.
\newblock {Monte Carlo} calculations of the radial distribution functions for a
  proton-electron plasma.
\newblock \emph{Australian Journal of Physics}, 18\penalty0 (2):\penalty0 119,
  1965.
\newblock \doi{10.1071/ph650119}.

\bibitem[Barp et~al.(2019)Barp, Briol, Duncan, Girolami, and
  Mackey]{BarpBriolDuncanEtAl2019Minimum}
Barp, A., Briol, F.-X., Duncan, A., Girolami, M., and Mackey, L.
\newblock Minimum {Stein} discrepancy estimators.
\newblock In \emph{Advances in Neural Information Processing Systems},
  volume~32, 2019.

\bibitem[Blei et~al.(2003)Blei, Ng, and Jordan]{BleNgJor03}
Blei, D., Ng, A., and Jordan, M.
\newblock Latent {D}irichlet allocation.
\newblock \emph{Journal of Machine Learning Research}, 3:\penalty0 993--1022,
  2003.

\bibitem[Bresler \& Nagaraj(2019)Bresler and Nagaraj]{BreNag19}
Bresler, G. and Nagaraj, D.
\newblock Stein's method for stationary distributions of {Markov} chains and
  application to {Ising} models.
\newblock \emph{Annals of Applied Probability}, 29, 2019.

\bibitem[Chwialkowski et~al.(2016)Chwialkowski, Strathmann, and
  Gretton]{chwialkowski_kernel_2016}
Chwialkowski, K., Strathmann, H., and Gretton, A.
\newblock A kernel test of goodness of fit.
\newblock In \emph{International conference on machine learning}, pp.\
  2606--2615. {PMLR}, 2016.

\bibitem[Cuturi(2011)]{Cuturi2011}
Cuturi, M.
\newblock Fast global alignment kernels.
\newblock In \emph{Proceedings of the 28th International Conference on
  International Conference on Machine Learning}, ICML'11, pp.\  929–936,
  Madison, WI, USA, 2011. Omnipress.
\newblock ISBN 9781450306195.

\bibitem[Cuturi et~al.(2007)Cuturi, Vert, Birkenes, and Matsui]{Cuturi_2007}
Cuturi, M., Vert, J.-P., Birkenes, O., and Matsui, T.
\newblock A kernel for time series based on global alignments.
\newblock 2007.
\newblock \doi{10.1109/icassp.2007.366260}.

\bibitem[Dehling \& Mikosch(1994)Dehling and Mikosch]{dehling_random_1994}
Dehling, H. and Mikosch, T.
\newblock Random quadratic forms and the bootstrap for {U}-statistics.
\newblock \emph{Journal of Multivariate Analysis}, 51\penalty0 (2):\penalty0
  392--413, 1994.
\newblock Publisher: Elsevier.

\bibitem[Fernandez et~al.(2020)Fernandez, Rivera, Xu, and
  Gretton]{fernandez_kernelized_2020}
Fernandez, T., Rivera, N., Xu, W., and Gretton, A.
\newblock Kernelized {Stein} discrepancy tests of goodness-of-fit for
  time-to-event data.
\newblock In \emph{Proceedings of the 37th International Conference on Machine
  Learning}, pp.\  3112--3122, 2020.

\bibitem[Gorham \& Mackey(2015)Gorham and Mackey]{gorham_measuring_2015}
Gorham, J. and Mackey, L.
\newblock Measuring sample quality with {Stein}'s method.
\newblock In \emph{Advances in Neural Information Processing Systems},
  volume~28, 2015.

\bibitem[Gorham \& Mackey(2017)Gorham and Mackey]{GorMac2017}
Gorham, J. and Mackey, L.
\newblock Measuring sample quality with kernels.
\newblock In \emph{{Proceedings of The 34th International Conference on Machine
  Learning}}, pp.\  1292--1301, 2017.

\bibitem[Gretton et~al.(2012)Gretton, Borgwardt, Rasch, Schölkopf, and
  Smola]{gretton_kernel_2012}
Gretton, A., Borgwardt, K.~M., Rasch, M.~J., Schölkopf, B., and Smola, A.
\newblock A kernel two-sample test.
\newblock \emph{The Journal of Machine Learning Research}, 13\penalty0
  (1):\penalty0 723--773, 2012.
\newblock Publisher: {JMLR}. org.

\bibitem[Haussler(1999)]{haussler_convolution_1999}
Haussler, D.
\newblock Convolution kernels on discrete structures, 1999.

\bibitem[Hodgkinson et~al.(2020)Hodgkinson, Salomone, and
  Roosta]{hodgkinson_reproducing_2020}
Hodgkinson, L., Salomone, R., and Roosta, F.
\newblock The reproducing stein kernel approach for post-hoc corrected
  sampling.
\newblock 2020.

\bibitem[Hoeffding(1948)]{hoeffding_class_1948}
Hoeffding, W.
\newblock A class of statistics with asymptotically normal distribution.
\newblock \emph{Ann. Math. Statist.}, 19\penalty0 (3):\penalty0 293--325,
  September 1948.

\bibitem[Kallenberg(2021)]{Kallenberg_2021}
Kallenberg, O.
\newblock \emph{Foundations of Modern Probability}.
\newblock Springer International Publishing, 2021.
\newblock \doi{10.1007/978-3-030-61871-1}.

\bibitem[Kanagawa et~al.(2023)Kanagawa, Jitkrittum, Mackey, Fukumizu, and
  Gretton]{kanagawa_kernel_2019}
Kanagawa, H., Jitkrittum, W., Mackey, L., Fukumizu, K., and Gretton, A.
\newblock A kernel {S}tein test for comparing latent variable models.
\newblock To appear in Journal of the Royal Statistical Society. Series B,
  (Statistical Methodology), 2023.

\bibitem[Király \& Oberhauser(2019)Király and
  Oberhauser]{kiraly_kernels_2019}
Király, F.~J. and Oberhauser, H.
\newblock Kernels for sequentially ordered data.
\newblock 20, 2019.
\newblock Publisher: Journal of Machine Learning Research.

\bibitem[Leslie \& Kuang(2004)Leslie and Kuang]{Leslie2004}
Leslie, C. and Kuang, R.
\newblock Fast string kernels using inexact matching for protein sequences.
\newblock \emph{Journal of Machine Learning Research}, 5:\penalty0 1435–1455,
  dec 2004.
\newblock ISSN 1532-4435.

\bibitem[Liu et~al.(2016)Liu, Lee, and Jordan]{liu_kernelized_2016}
Liu, Q., Lee, J., and Jordan, M.
\newblock A kernelized {Stein} discrepancy for goodness-of-fit tests.
\newblock In \emph{International conference on machine learning}, pp.\
  276--284. {PMLR}, 2016.

\bibitem[Lloyd \& Ghahramani(2015)Lloyd and Ghahramani]{LloGha15}
Lloyd, J. and Ghahramani, Z.
\newblock Statistical model criticism using kernel two sample tests.
\newblock In \emph{Advances in Neural Information Processing Systems}, pp.\
  829--837, 2015.

\bibitem[Lodhi et~al.(2002)Lodhi, Saunders, Shawe-Taylor, Cristianini, and
  Watkins]{lodhi_text_2002}
Lodhi, H., Saunders, C., Shawe-Taylor, J., Cristianini, N., and Watkins, C.
\newblock Text classification using string kernels.
\newblock 2:\penalty0 419--444, 2002.

\bibitem[Oates et~al.(2016)Oates, Girolami, and Chopin]{Oates_2016}
Oates, C.~J., Girolami, M., and Chopin, N.
\newblock Control functionals for {Monte Carlo} integration.
\newblock \emph{Journal of the Royal Statistical Society: Series B (Statistical
  Methodology)}, 79\penalty0 (3):\penalty0 695--718, 2016.
\newblock \doi{10.1111/rssb.12185}.

\bibitem[Power \& Goldman(2019)Power and Goldman]{power_accelerated_2019}
Power, S. and Goldman, J.~V.
\newblock Accelerated sampling on discrete spaces with non-reversible markov
  processes.
\newblock 2019.

\bibitem[Radford et~al.(2019)Radford, Wu, Child, Luan, Amodei, and
  Sutskever]{radford2019language}
Radford, A., Wu, J., Child, R., Luan, D., Amodei, D., and Sutskever, I.
\newblock Language models are unsupervised multitask learners.
\newblock 2019.

\bibitem[Reinert \& Ross(2019)Reinert and Ross]{ReiRos19}
Reinert, G. and Ross, N.
\newblock Approximating stationary distributions of fast mixing {Glauber}
  dynamics, with applications to exponential random graphs.
\newblock \emph{Annals of Applied Probability}, 29, 2019.

\bibitem[Serfling(2009)]{Ser2009}
Serfling, R.~J.
\newblock \emph{Approximation theorems of mathematical statistics}.
\newblock John Wiley \& Sons, 2009.

\bibitem[Shi et~al.(2022)Shi, Zhou, Hwang, Titsias, and
  Mackey]{shi_gradient_2022}
Shi, J., Zhou, Y., Hwang, J., Titsias, M.~K., and Mackey, L.~W.
\newblock Gradient estimation with discrete {Stein} operators.
\newblock In \emph{Advances in Neural Information Processing Systems},
  volume~35, 2022.

\bibitem[Sriperumbudur et~al.(2011)Sriperumbudur, Fukumizu, and
  Lanckriet]{SriFukLan2011}
Sriperumbudur, B.~K., Fukumizu, K., and Lanckriet, G. R.~G.
\newblock Universality, characteristic kernels and {RKHS} embedding of
  measures.
\newblock \emph{Journal of Machine Learning Research}, 12:\penalty0 2389--2410,
  2011.

\bibitem[Stein(1972)]{stein_bound_1972}
Stein, C.~M.
\newblock A bound for the error in the normal approximation to the distribution
  of a sum of dependent random variables.
\newblock 1972.

\bibitem[Steinwart \& Christmann(2008)Steinwart and
  Christmann]{steinwart_support_2008}
Steinwart, I. and Christmann, A.
\newblock \emph{Support Vector Machines}.
\newblock Information Science and Statistics. Springer, 2008 edition, 2008.

\bibitem[Wynne et~al.(2022)Wynne, Kasprzak, and Duncan]{Wynne2022}
Wynne, G., Kasprzak, M., and Duncan, A.~B.
\newblock A spectral representation of kernel {Stein} discrepancy with
  application to goodness-of-fit tests for measures on infinite dimensional
  hilbert spaces.
\newblock 2022.
\newblock \doi{https://doi.org/10.48550/arXiv.2206.04552}.

\bibitem[Xu \& Matsuda(2020)Xu and Matsuda]{xu_2020}
Xu, W. and Matsuda, T.
\newblock A {Stein} goodness-of-fit test for directional distributions.
\newblock In \emph{Proceedings of the Twenty Third International Conference on
  Artificial Intelligence and Statistics}, pp.\  320--330, 2020.

\bibitem[Yang et~al.(2018)Yang, Liu, Rao, and Neville]{yang_goodness--fit_2018}
Yang, J., Liu, Q., Rao, V.~A., and Neville, J.
\newblock Goodness-of-fit testing for discrete distributions via {Stein}
  discrepancy.
\newblock In \emph{{ICML}}, 2018.

\bibitem[Yang et~al.(2019)Yang, Rao, and Neville]{yang_2019}
Yang, J., Rao, V., and Neville, J.
\newblock A {Stein–Papangelou} goodness-of-fit test for point processes.
\newblock In \emph{Proceedings of the Twenty-Second International Conference on
  Artificial Intelligence and Statistics}, pp.\  226--235, 2019.

\bibitem[Zanella(2019)]{zanella_informed_2019}
Zanella, G.
\newblock Informed proposals for local {MCMC} in discrete spaces.
\newblock 115:\penalty0 852 -- 865, 2019.

\end{thebibliography}
\bibliographystyle{icml2023}

\newpage
\appendix
\onecolumn
\section{Experimental details}
We provide further details for the experiments that we conducted.

\subsection{Likelihood Ratio Test}

The LR baselines use the likelihood ratio statistic:
\begin{align}
    T^{\text{LR}}_\disSampleCount\of{\disSamples} &\coloneqq 2 \of{
        \sup_{\pmft\in H_0\cup H_1} \log \pmft\of\disSamples
        - \sup_{\pmft\in H_0} \log \pmft\of\disSamples
    },
\end{align}
where $H_0$ and $H_1$ are the null and alternative hypotheses,
respectively.
In our experiments, we use a simple null hypothesis for the likelihood
ratio statistic.
The alternative used in the oracle baseline is a simple hypothesis
containing only the (normally unknown) ground truth.
The alternative used in the MC baseline is a composite hypothesis
consisting of all Markov chains of some order,
with the order depending on the experiment.
In experiments where we use an alternative consisting of
first-order Markov chains,
the alternative consists of all initial distributions
over the alphabet together with all transition kernels
through the alphabet, with an additional stopping point:
\begin{align}
    H_1 &\coloneqq \simplex{\alphabet} \times
    \of{\simplex{\alphabet\cup\{\text{stop}\}}}^{\alphabet}, \\
    \simplex{\alphabet} &\coloneqq \left\{
    (\theta_i)_{i\in\alphabet} :
    \sum_{i\in\alphabet} \theta_i = 1
    \text{ and }\theta_i \ge 0 \,\text{ for any } i\in\alphabet
    \right\}.
\end{align}
In experiments where we use an alternative consisting of
second-order Markov chains,
the alternative consists of the analogous
collection of second-order kernels:
\begin{align}
    H_1 &\coloneqq
    \simplex\alphabet \times
    \of{\simplex{\alphabet\cup\{\text{stop}\}}}^\alphabet \times
    \of{\simplex{\alphabet\cup\{\text{stop}\}}}^{\alphabet\times\alphabet}.
\end{align}

\subsection{Further Details Regarding the Language Model Experiment in \cref{sec:additional-exp-conditional}}\label{sec:toy-experiment-details}
In \cref{sec:additional-exp-conditional}, we consider the problem of evaluating a generative model with an output constraint. 
Although the support of the distribution is restricted to the set of question sentences, our setting still covers this scenario 
because we may construct a question sentence from any sequence by adding the suffix '?' (i.e., the set of all sequences is bijective to that of all question sequences). 
In reality, however, the assumption that the model has positive probability everywhere is unlikely to hold; e.g., some tokens never appear in questions (our experiment limits possible sequences with prompts). 
This assumption is placed to ensure that the formalism is well-defined; in practice, to follow the formalism, we may assume that the model has extremely small probabilities that do not affect the model's output effectively. 
That said, it is more useful to consider a neighborhood structure that spans the model's support rather than the entire sequence space. 
As described below, our neighborhood design excludes any points which are zero-probability under the model. 

The model is configured using \emph{top-$k$ filtering} with $k=16$. The sampling procedure is as follows: When the model samples the next token, $x_{n+1}$,
for an already-sampled prefix $s=(x_1,\ldots,x_{n})$, only the most likely $k$ tokens from the alphabet are considered.
That is, we sort the alphabet $\alphabet$ using the conditional model density
$\cprob{x_{n+1}=\bullet}{x_1,\ldots,x_n}$, and keep only the $k$ elements which have the largest probability
using this ordering.
Then, we renormalize the probabilities of these $k$ points and sample the next token from the resulting distribution.

As mentioned in the main body, we use point-dependent insertion and substitution sets. 
Specifically, for a given sequence $s=(x_1,\ldots,x_n,'?')$, the substitution neighborhood $\mathcal{R}_{x_{(1:n)},n}$ consists of all sequences of the form $s'=(x_1,\ldots,x_{n}', '?')$, where $x_{n}'$ is chosen from any of the top-$k$ next tokens after the prefix $s_{(1:n-1)}$.
Similarly, the insertion neighborhood $\mathcal{I}_{x_{(1:n)},n}$ consists of all top-$k$ sequences $s'=(x_1,\ldots,x_n,x_{n+1}', '?')$, where $x_{n+1}'$ is any of the top-$k$ next tokens after the prefix $s_{(1:n)}$. 
The deletion neighborhood consists of the single sequence $s'=(x_1,\ldots,x_{n-1},'?')$.
Since an observed point must have non-zero probability (otherwise we can immediately reject the null hypothesis), $x_{n}$ must itself be one of the top-$k$ next tokens after the prefix $s_{(1:n-1)}$
Hence, substitutions are symmetric.
Insertions are similarly symmetric. 

This prefix-dependent neighborhood generalizes the approach described in \Cref{sec:choice-of-neighbourhood-graph}, where we used a subset of the alphabet as a symbol neighborhood $\mathcal{N}_{\mathrm{rep}}(s)=\mathcal{N}_{\mathrm{rep}}(x_{n})\subsetneq \mathcal{S}$. Our generalization is that the substitution neighborhood depends not on the symbol being replaced, but on the point as a whole via the prefix. Similarly, the insertion neighborhood also depends on the point as a whole.

The neighborhood graph used in the $\mathrm{ZS}'$ variant is not strongly connected, rendering the resulting test inconsistent.
The neighborhood graph used in the $\mathrm{ZS}$ variant is not guaranteed to be strongly connected.
This is because the neighborhood graph that we use satisfies two conditions: the edges consist only of transitions with a bounded edit distance (all transitions have edit distance 1), and transitions to
zero-probability sequences are excluded from the graph.
Depending on the model structure, it may then be the case that the neighborhood graph is disconnected.
For example, the sequences \texttt{Where is my pigeon?} and \texttt{Where is the cat?} may be in the support
of the model.
A path in the neighborhood graph used in the $\mathrm{ZS}$ variant would have to pass through the sequences
\texttt{Where is my?} and \texttt{Where is the?}.
We would not expect the these rather unnatural sentences to be in the support set of the model when top-$k$ filtering is used,
and in that case the necessary transitions would be excluded from the neighborhood graph.
This would cause the graph to be disconnected.

We use the pretrained model from Hugging Face (\url{https://huggingface.co/gpt2}) to conduct the experiment. 

\subsection{Details of Testing Scenarios}\label{sec:testing-scenarios}

In several experiments, we referred to a collection of 12 synthetic
testing scenarios to evaluate the test power.
We provide the details of these scenarios.
The scenarios are intended to be dissimilar to each other
in form and difficulty,
although all model distributions and ground truths are Markov
chains of some order.

Some of the distributions described here 
do not have support on the entire sequence space.
In our earlier discussion, we required support everywhere
as a technical condition relating to the connectivity of the Zanella process.
To ensure that the probability of all sequences is positive,
we introduce low-probability ``restart'' events into all of our Markov chains
that resample uniformly from the alphabet with low probability,
rather than sampling from the specified transition kernel.
The probability of these events is small, $\varepsilon\approx 10^{-3}$.
We do not mention this further, in order to simplify our discussion.

\paragraph{Binary i.i.d.\ sequences}
We draw sequences of i.i.d.\ Bernoulli variables.
The model and ground truth differ in the parameter
of the Bernoulli distribution.

Alphabet: $\{0,1\}$.

Model distribution:
Length follows a Poisson distribution with mean 20.
Contents of the sequence are i.i.d.\ Bernoulli with $p=0.6$.

Ground truth:
Length follows a Poisson distribution with mean 20.
Contents of the sequence are i.i.d.\ Bernoulli with $p=0.4$.

Sample size: 10.

\paragraph{Binary sequences: misspecified Markov order}
We simulate the case where we are mistaken about the Markov order
of the data.
The model consists of i.i.d.\ Bernoulli variables,
while the ground truth follows a simple first-order Markov chain.

Alphabet: $\{0,1\}$.

Model distribution:
Length follows a geometric distribution with mean 20.
Contents of the sequence are i.i.d.\ Bernoulli with $p=0.6$.

Ground truth:
First-order Markov chain.
Initial distribution uniform on \{0, 1\}.
The transition kernel is 
\begin{align}
    \cprob{\text{stop}}{X_t=x} &= 1/\lambda, \\
    \cprob{X_{t+1}=y}{X_t=x} &= \of{1-1/\lambda} \delta_x(y),
\end{align}
with $\lambda=20$.
This gives sequences alternating between $\{0,1\}$.

Sample size: 30.

\paragraph{Random walk I}
We take a random walk through a cyclic alphabet
and perturb it by letting the ground truth hold, or ``skip a step'',
with low probability.

Alphabet: $\alphabet=\{1,\ldots,8\}$
with a cyclic structure.

Model distribution:
Random walk through $\alphabet$ according to the cyclic structure.
Start uniformly on $\alphabet$.
After each step, stop with probability $1/\lambda$, with $\lambda=8$.

Ground truth:
Random walk with holding.
Start uniformly on $\alphabet$.
At each step, hold or ``skip a step'',
with probability $p=0.2$, letting $X_{t+1}=X_t$.
Otherwise, sample a step from $\{-1,+1\}$ uniformly.
After each step, stop with probability $1/\lambda$, with $\lambda=8$.

Sample size: 30.

\paragraph{Random walk II}
The same scenario, but with few long sequences instead
of many short sequences.
We only let the ground truth hold in a subset of states,
which reduces the degree of perturbation and makes the problem
more challenging.

Alphabet: $\alphabet=\{1,\ldots,30\}$
with a cyclic structure.

Model distribution:
Same as in \texttt{Random walk I} with $\lambda=30$.

Ground truth:
Same as in \texttt{Random walk I} but with $\lambda=30$
and sampling the step from $\{-1,0,+1\}$ according to:
\begin{align}
    \cprob{X_{t+1}-X_t = \Delta}{X_t = x} &=
    \begin{cases}
        p, &\text{if } \Delta=0, x\le i_{\text{max}}, \\
        \frac{1-p}{2}, &\text{if } \Delta\in\{-1,+1\}, x\le i_{\text{max}}, \\
        \frac{1}{2}, &\text{if } \Delta\in\{-1,+1\}, x>i_{\text{max}}, \\
        0, &\text{otherwise,}
    \end{cases}
\end{align}
with $i_\text{max}=8$ and $p=0.2$.

Sample size: 8.

\paragraph{Random walk III}
We introduce memory into the random walk by correlating
or anti-correlating successive steps,
which gives us a simple second-order MC.

Alphabet: $\alphabet=\{1,\ldots,10\}$
with a cyclic structure.

Model distribution:
Same as in \texttt{Random walk I}, but we modify the transition
kernel to correlate successive steps.
Specifically, letting $\delta=0.95$,
we sample the step from $\{-1,0,+1\}$ according to:
\begin{align}
    \cprob{X_{t+2}-X_{t+1}=d}{X_{t+1}, X_t} &=
    \begin{cases}
        \delta, &\text{if } d=X_{t+1}-X_t, |d|=1, \\
        1-\delta, &\text{if } d=X_t-X_{t+1}, |d|=1, \\
        \frac{1}{2}, &\text{if } |d|=1<|X_t-X_{t+1}|, \\
        0, &\text{otherwise.}
    \end{cases} \\
\end{align}
We have to treat the case where $|X_t-X_{t+1}|>1$ due to
the low probability restart event mentioned earlier.

Ground truth:
Same as model distribution but with $\delta=0.05$.

Sample size: 30.

\paragraph{Random walk IV}
Same as in \texttt{Random walk III}
but with $\lambda=30$ and sample size 8.
That is, few long sequences instead of many short ones.

\paragraph{Random second-order MC I}
We define a distribution over second-order MC and sample two random ones.
The stopping probability is constant so that we can control
the length independently of the perturbation of the contents.
We sample many short sequences.

Alphabet: $\alphabet=\{1,\ldots,10\}$.

Model distribution:
We sample a second-order Markov chain
by sampling the initial distribution and transition kernels
from Dirichlet distributions with concentration parameter $\alpha=1$.
At each step, the chain stops with probability $1/\lambda$
where $\lambda=8$.

Ground truth:
Same as model distribution, but the Markov chain
is sampled from a different random number generator (RNG) seed
in order to give a different MC.

Sample size: 30.

\paragraph{Random second-order MC II}
Same as in \texttt{Random second-order MC I}
but with $\lambda=20$ and sample size 8.
That is, we sample a small number of long sequences.

\paragraph{Random second-order MC III}
Same as in \texttt{Random second-order MC I}
but with $\lambda=8$ and sample size 8.
That is, we sample a small number of short sequences.

\paragraph{Random MC with varied initial distribution I}
We sample a single random first-order MC for the model
and ground truth, and set a fixed initial distribution
which we perturb.
This poses a hard problem,
as most of the observed data relates to transitions that come
from an unperturbed transition kernel.

Alphabet: $\alphabet=\{1,\ldots,10\}$.

Model distribution:
We sample a first-order Markov chain
by sampling the transition kernel
from a Dirichlet distribution with concentration parameter $\alpha=1$.
The initial distribution is uniform over $\alphabet$.
At each step, the chain stops with probability $1/\lambda$
where $\lambda=8$.

Ground truth:
Same as model distribution, with the same RNG seed.
in order to give the same transition kernel.
The initial distribution is an equal mixture
of $\distUnif{\alphabet}$
and $\distUnif{\{1,2\}}$.

Sample size: 30.

\paragraph{Random MC with varied initial distribution II}
Same as in \texttt{Random MC with varied initial distribution I}
but with $\lambda=20$ and sample size 8.
This problem is significantly harder, as we observe 8 rather than
30 instances of the initial distribution, and the longer
sequences do not provide any additional evidence.

\paragraph{Random MC with varied length distribution}
We vary the distribution over lengths while keeping
the content distribution the same.

Alphabet: $\alphabet=\{1,\ldots,10\}$.

Model distribution:
We sample a first-order Markov chain
by sampling the transition kernel
from a Dirichlet distribution with concentration parameter $\alpha=1$.
The initial distribution is uniform over $\alphabet$.
At each step, the chain stops with probability $1/\lambda$
where $\lambda=8$.

Ground truth:
We use the same initial distribution
and transition kernel as in the model.
At each step, the chain stops with probability $1/\lambda$
where $\lambda=20$.

Sample size: 30.

\subsection{Results of Individual Testing Scenarios}\label{sec:scenarios-table}
We report the individual per-scenario rejection rates for an earlier experiment,
for which we reported the average rate across scenarios in \Cref{fig:simple-neigh-size}.
The per-scenario rates are shown in \Cref{fig:simple-neigh-size-perscenario}.

Two of the scenarios are designed to highlight the failure of the inconsistent
test variants. These are \emph{Binary sequences: True distribution is not i.i.d.\ sequence},
which shows the failure of the inconsistent CSK kernel; and
\emph{Random MC w/ varied length dist}, which shows the failure of
the inconsistent variant $\mathrm{ZS}'$.

\begin{sidewaystable}
\tiny
\begin{tabular}{lrrrrrrrrrrrrrrrrrr}
Test & \multicolumn{6}{r}{CSK $\mathrm{ZS}$} & \multicolumn{6}{r}{CSK $\mathrm{ZS}'$} & \multicolumn{5}{r}{CSK $\mathrm{ZS}''$} & CSK MMD \\
J & 1 & 3 & 5 & 7 & 9 & $\infty$ & 1 & 3 & 5 & 7 & 9 & $\infty$ & 1 & 3 & 5 & 7 & 9 &  \\
Scenario &  &  &  &  &  &  &  &  &  &  &  &  &  &  &  &  &  &  \\
Binary sequences: Few long i.i.d. sequences & 0.237 & 0.415 & 0.715 & 0.870 & 0.932 & 1.000 & 0.242 & 0.507 & 0.792 & 0.902 & 0.960 & 1.000 & 0.225 & 0.375 & 0.623 & 0.690 & 0.810 & 0.995 \\
Binary sequences: True distribution is not i.i.d. sequence & 0.075 & 0.022 & 0.050 & 0.030 & 0.030 & 0.040 & 0.033 & 0.007 & 0.013 & 0.018 & 0.013 & 0.005 & 0.065 & 0.048 & 0.050 & 0.080 & 0.090 & 0.013 \\
Random 2nd-order MC: Few long sequences & 0.040 & 0.158 & 0.282 & 0.438 & 0.500 & 0.907 & 0.075 & 0.172 & 0.335 & 0.420 & 0.517 & 0.917 & 0.045 & 0.180 & 0.253 & 0.352 & 0.472 & 0.670 \\
Random 2nd-order MC: Few short sequences & 0.065 & 0.117 & 0.225 & 0.280 & 0.357 & 0.477 & 0.068 & 0.155 & 0.207 & 0.315 & 0.360 & 0.443 & 0.033 & 0.135 & 0.255 & 0.297 & 0.375 & 0.125 \\
Random 2nd-order MC: Many short sequences & 0.177 & 0.420 & 0.720 & 0.835 & 0.907 & 0.973 & 0.175 & 0.490 & 0.760 & 0.887 & 0.910 & 0.968 & 0.080 & 0.422 & 0.710 & 0.877 & 0.935 & 0.970 \\
Random MC w/ varied initial dist: Few long sequences & 0.052 & 0.065 & 0.060 & 0.068 & 0.048 & 0.043 & 0.043 & 0.043 & 0.055 & 0.028 & 0.048 & 0.040 & 0.070 & 0.065 & 0.060 & 0.055 & 0.062 & 0.100 \\
Random MC w/ varied initial dist: Many short sequences & 0.075 & 0.052 & 0.102 & 0.048 & 0.060 & 0.050 & 0.055 & 0.080 & 0.090 & 0.068 & 0.150 & 0.105 & 0.068 & 0.062 & 0.065 & 0.098 & 0.075 & 0.125 \\
Random MC w/ varied length dist & 0.033 & 0.018 & 0.013 & 0.018 & 0.025 & 0.035 & 0.025 & 0.007 & 0.028 & 0.045 & 0.030 & 0.025 & 0.013 & 0.000 & 0.010 & 0.018 & 0.010 & 0.690 \\
Random walk with memory: Few long sequences & 0.220 & 0.323 & 0.487 & 0.570 & 0.580 & 0.978 & 0.395 & 0.465 & 0.517 & 0.580 & 0.655 & 0.958 & 0.085 & 0.060 & 0.133 & 0.223 & 0.263 & 1.000 \\
Random walk with memory: Many short sequences & 0.525 & 0.710 & 0.892 & 0.943 & 0.968 & 0.995 & 0.693 & 0.762 & 0.907 & 0.943 & 0.980 & 0.998 & 0.135 & 0.207 & 0.393 & 0.500 & 0.733 & 1.000 \\
Random walk: Few long sequences & 0.130 & 0.383 & 0.740 & 0.810 & 0.880 & 1.000 & 0.147 & 0.405 & 0.603 & 0.755 & 0.800 & 1.000 & 0.030 & 0.417 & 0.652 & 0.838 & 0.925 & 0.595 \\
Random walk: Many short sequences & 0.438 & 0.912 & 0.978 & 0.993 & 1.000 & 1.000 & 0.647 & 0.818 & 0.963 & 0.995 & 0.998 & 0.998 & 0.072 & 0.810 & 0.970 & 0.993 & 1.000 & 0.618 \\
\end{tabular}

\begin{tabular}{lrrrrrrrrrrrrrrrrrr}
Test & \multicolumn{6}{r}{Hamming $\mathrm{ZS}$} & \multicolumn{6}{r}{Hamming $\mathrm{ZS}'$} & \multicolumn{5}{r}{Hamming $\mathrm{ZS}''$} & Hamming MMD \\
J & 1 & 3 & 5 & 7 & 9 & $\infty$ & 1 & 3 & 5 & 7 & 9 & $\infty$ & 1 & 3 & 5 & 7 & 9 &  \\
Scenario &  &  &  &  &  &  &  &  &  &  &  &  &  &  &  &  &  &  \\
Binary sequences: Few long i.i.d. sequences & 0.030 & 0.065 & 0.070 & 0.043 & 0.040 & 0.100 & 0.100 & 0.092 & 0.133 & 0.220 & 0.215 & 0.195 & 0.033 & 0.037 & 0.040 & 0.052 & 0.037 & 0.060 \\
Binary sequences: True distribution is not i.i.d. sequence & 0.055 & 0.117 & 0.170 & 0.215 & 0.180 & 0.152 & 0.133 & 0.347 & 0.492 & 0.603 & 0.745 & 0.815 & 0.060 & 0.120 & 0.138 & 0.177 & 0.220 & 0.087 \\
Random 2nd-order MC: Few long sequences & 0.058 & 0.225 & 0.225 & 0.263 & 0.207 & 0.237 & 0.062 & 0.085 & 0.117 & 0.120 & 0.128 & 0.113 & 0.035 & 0.140 & 0.220 & 0.190 & 0.237 & 0.050 \\
Random 2nd-order MC: Few short sequences & 0.077 & 0.177 & 0.190 & 0.280 & 0.205 & 0.210 & 0.077 & 0.142 & 0.090 & 0.150 & 0.135 & 0.142 & 0.075 & 0.160 & 0.228 & 0.237 & 0.223 & 0.052 \\
Random 2nd-order MC: Many short sequences & 0.115 & 0.188 & 0.233 & 0.233 & 0.315 & 0.287 & 0.205 & 0.245 & 0.273 & 0.285 & 0.280 & 0.247 & 0.052 & 0.217 & 0.205 & 0.265 & 0.273 & 0.080 \\
Random MC w/ varied initial dist: Few long sequences & 0.065 & 0.022 & 0.060 & 0.070 & 0.062 & 0.035 & 0.037 & 0.058 & 0.045 & 0.077 & 0.062 & 0.090 & 0.048 & 0.062 & 0.070 & 0.052 & 0.040 & 0.043 \\
Random MC w/ varied initial dist: Many short sequences & 0.140 & 0.085 & 0.030 & 0.070 & 0.062 & 0.065 & 0.163 & 0.210 & 0.172 & 0.163 & 0.217 & 0.230 & 0.040 & 0.060 & 0.068 & 0.068 & 0.060 & 0.070 \\
Random MC w/ varied length dist & 0.005 & 0.015 & 0.025 & 0.037 & 0.033 & 0.225 & 0.015 & 0.013 & 0.005 & 0.010 & 0.003 & 0.000 & 0.013 & 0.007 & 0.010 & 0.030 & 0.030 & 0.275 \\
Random walk with memory: Few long sequences & 0.080 & 0.060 & 0.065 & 0.035 & 0.068 & 0.085 & 0.068 & 0.113 & 0.077 & 0.090 & 0.045 & 0.060 & 0.052 & 0.037 & 0.070 & 0.052 & 0.075 & 0.077 \\
Random walk with memory: Many short sequences & 0.030 & 0.048 & 0.125 & 0.075 & 0.080 & 0.077 & 0.033 & 0.065 & 0.062 & 0.083 & 0.050 & 0.075 & 0.083 & 0.087 & 0.102 & 0.080 & 0.110 & 0.098 \\
Random walk: Few long sequences & 0.043 & 0.072 & 0.113 & 0.138 & 0.175 & 0.225 & 0.068 & 0.055 & 0.087 & 0.043 & 0.065 & 0.075 & 0.080 & 0.102 & 0.105 & 0.147 & 0.122 & 0.068 \\
Random walk: Many short sequences & 0.050 & 0.152 & 0.155 & 0.182 & 0.150 & 0.215 & 0.043 & 0.110 & 0.115 & 0.102 & 0.128 & 0.077 & 0.080 & 0.170 & 0.160 & 0.170 & 0.245 & 0.060 \\
\end{tabular}

\caption{%
        Estimated rejection rate of three different families of Zanella-Stein test with varying neighborhood size
        and underlying kernel.
        We evaluate the tests on 12 different synthetic testing scenarios,
        evaluating each test multiple times on independently sampled synthetic datasets.
        To estimate the per-scenario rejection rate,
        we report an average across the test evaluations for each scenario.
        The data shown here is averaged across scenarios in \Cref{fig:simple-neigh-size} above.
}
\label{fig:simple-neigh-size-perscenario}
\end{sidewaystable}

\newpage
\section{Additional Experiments}
\subsection{Balancing Function}
We conduct experiments to show the relative performance of the Barker
balancing function $t\mapsto t/(1+t)$ against the minimum probability flow (MPF) $t\mapsto \sqrt{t}$. 
\Cref{tab:simple-exp-bal-fun} shows performance of a KSD on the 12 testing scenarios
detailed in \cref{sec:testing-scenarios},
comparing the two balancing functions.
\Cref{fig:exp-barker-vs-mpf} shows performance of a KSD on the MRF experiment,
comparing the two balancing functions against various baselines.
We observe that the MPF operator tends to yield higher power than the Barker operator.
A drawback of the MPF operator is that the ratio in \Cref{eqn:steinop} can be numerically unstable when $p(x)$ is extremely small.
The Barker operator circumvents this issue as the ratio becomes $p(y)/\{p(x)+p(y)\}$ \Citep[][also mentions this point]{shi_gradient_2022}.

\begin{table}[h]
    \centering
    \pgfkeys{/pgf/number format/.cd,fixed,fixed zerofill,precision=2,skip 0.=true}
    \begin{tabular}{l|r|r}
        ~
        & Barker
        & MPF
        \\
        \hline
        Binary i.i.d.\ sequences
        & 0.88
        & 0.80
        \\
        Binary sequences: misspecified Markov order
        & 0.01
        & 0.00
        \\
        Random walk I
        & 0.98
        & 1.00
        \\
        Random walk II
        & 0.69
        & 0.82
        \\
        Random walk III
        & 0.93
        & 0.96
        \\
        Random walk IV
        & 0.54
        & 0.66
        \\
        Random second-order MC I
        & 0.84
        & 0.90
        \\
        Random second-order MC II
        & 0.38
        & 0.53
        \\
        Random second-order MC III
        & 0.28
        & 0.45
        \\
        Random MC with varied initial distribution I
        & 0.07
        & 0.05
        \\
        Random MC with varied initial distribution II
        & 0.05
        & 0.09
        \\
        Random MC with varied length distribution
        & 0.05
        & 0.02
    \end{tabular}
    \caption{%
        Estimated test power of the Zanella-Stein kernel goodness-of-fit test,
        constructed from the CSK kernel using two different balancing functions.
        We choose either the Barker or the minimum probability flow (MPF) function.
    }
    \label{tab:simple-exp-bal-fun}
\end{table}

\begin{figure}[h]
    \centering
    \begin{tikzpicture}[scale=0.8]
        \begin{axis}
            [
            xlabel=Concentration parameter $\theta$ of ground truth,
            ylabel=Estimated rejection rate,
            ymin=0, ymax=1,
            ytick={0,0.2,0.4,0.6,0.8,1},
            enlargelimits=0.1,
            legend pos=outer north east,
            legend cell align=left,
            cycle list name=linecyclelist,
            every axis plot/.append style={
                line width=1.5pt,
            },
            ]
            \addplot table[
            col sep=comma,
            x={sample_concentration_param},
            y={Stein-Wild(k=CSK: op=__mathrm{ZS}_{infty:b}_)},
            ]{figures/exp_simple_mrf_rejection_rate_by_sample_concentration_param.csv};
            \addlegendentry{KSD Barker};
            \addplot table[
            col sep=comma,
            x={sample_concentration_param},
            y={Stein-Wild(k=CSK: op=__mathrm{ZS}_{infty:mpf}_)},
            ]{figures/exp_simple_mrf_rejection_rate_by_sample_concentration_param.csv};
            \addlegendentry{KSD MPF};
            \addplot table[
            col sep=comma,
            x={sample_concentration_param},
            y={MMD-Wild(k=CSK: n=same)},
            ]{figures/exp_simple_mrf_rejection_rate_by_sample_concentration_param.csv};
            \addlegendentry{MMD};
            \addplot table[
            col sep=comma,
            x={sample_concentration_param},
            y={LR(oracle)},
            ]{figures/exp_simple_mrf_rejection_rate_by_sample_concentration_param.csv};
            \addlegendentry{LR oracle};
            \addplot[dashed,no markers] coordinates{ (0.75, 0.05) (1.25, 0.05) };
            \addlegendentry{Level $\alpha=0.05$};
        \end{axis}
    \end{tikzpicture}
    \caption{%
        Performance comparison of two choices of balancing function
        in the construction of the KSD:
        Barker against minimum probability flow.
    }
    \label{fig:exp-barker-vs-mpf}
\end{figure}
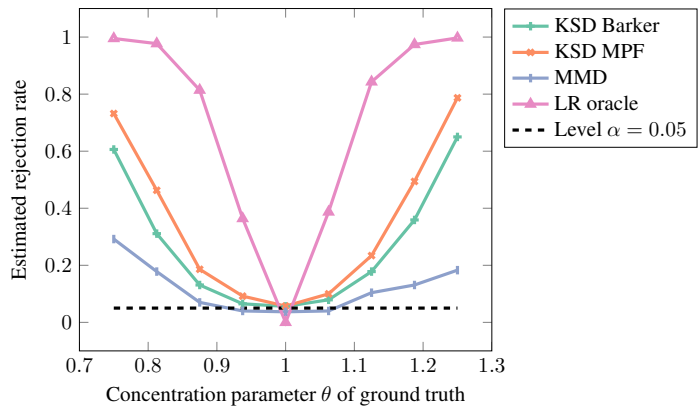

\newpage
\subsection{Larger Neighborhoods}
\begin{figure}[h]
    \centering
    \begin{tikzpicture}[scale=0.8]
        \begin{axis}
            [
            xlabel=Concentration parameter $\theta$ of ground truth,
            ylabel=Estimated rejection rate,
            ymin=0, ymax=1,
            ytick={0,0.2,0.4,0.6,0.8,1},
            enlargelimits=0.1,
            legend pos=outer north east,
            legend cell align=left,
            cycle list name=linecyclelist,
            every axis plot/.append style={
                line width=1.5pt,
            },
            ]
            \addplot table[
            col sep=comma,
            x={sample_concentration_param},
            y={Stein-Wild(k=CSK: op=__mathrm{ZS}_{infty:b}_)},
            ]{figures/exp_simple_mrf_rejection_rate_by_sample_concentration_param.csv};
            \addlegendentry{KSD, single edit};
            \addplot table[
            col sep=comma,
            x={sample_concentration_param},
            y={Stein-Wild(k=CSK: op=__mathrm{ZS}_{d=2:b}_)},
            ]{figures/exp_simple_mrf_rejection_rate_by_sample_concentration_param.csv};
            \addlegendentry{KSD, double edit};
            \addplot table[
            col sep=comma,
            x={sample_concentration_param},
            y={MMD-Wild(k=CSK: n=same)},
            ]{figures/exp_simple_mrf_rejection_rate_by_sample_concentration_param.csv};
            \addlegendentry{MMD};
            \addplot table[
            col sep=comma,
            x={sample_concentration_param},
            y={LR(oracle)},
            ]{figures/exp_simple_mrf_rejection_rate_by_sample_concentration_param.csv};
            \addlegendentry{LR oracle};
            \addplot[dashed,no markers] coordinates{ (0.75, 0.05) (1.25, 0.05) };
            \addlegendentry{Level $\alpha=0.05$};
        \end{axis}
    \end{tikzpicture}
    \caption{%
        Performance comparison of the single-edit neighbourhood test
        versus a variant which allows for double-edits.
    }
    \label{fig:exp-larger-edit-distance}
\end{figure}
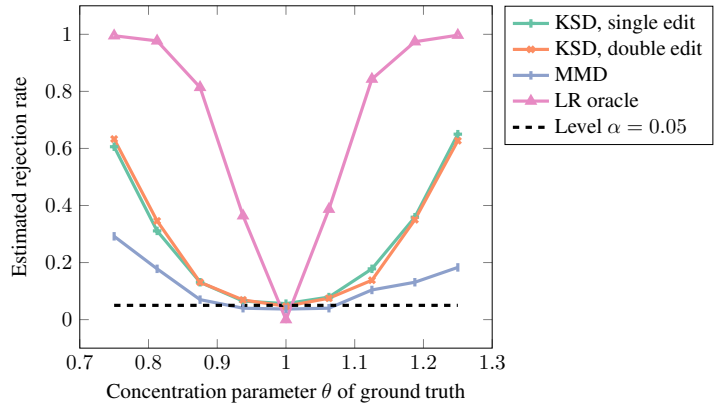

In \Cref{sec:method}, we considered a neighborhood structure where we modified one element of the sequence. 
Here, we conduct an experiment to see the effect of using a larger edit distance. 
We use the MRF experiment in \Cref{sec:mrf-experiment} as a benchmark 
and compare the original KSD used in \Cref{sec:mrf-experiment} against a KSD that allows for larger edit distance.
The neighborhood for the latter KSD is extended by adding double-substitution,
that is, substitution in two locations.
\Cref{fig:exp-larger-edit-distance} shows the result, where the oracle and MMD baselines are included as a reference.
We can see that test sensitivity is at best improved by a negligible amount in one region,
while similarly it worsens in the other region.
For this problem, expanding the neighborhood as above does not improve test sensitivity despite a higher computational cost. 

\end{document}